\documentclass[10pt,twocolumn,letterpaper]{article}

\usepackage{cvpr}              % To produce the CAMERA-READY version
\usepackage{graphicx}
\usepackage{amsmath}
\usepackage{amssymb}
\usepackage{booktabs}
\usepackage{caption}
\usepackage{colortbl}
\usepackage[table]{xcolor}
\usepackage{makecell}
\usepackage{float}
\usepackage{paralist}
\usepackage{xcolor}
\usepackage{threeparttable}
\usepackage{lipsum}

\makeatletter
\def\@xfootnote[#1]{%
  \protected@xdef\@thefnmark{#1}%
  \@footnotemark\@footnotetext}
\makeatother
\usepackage[pagebackref,breaklinks,colorlinks]{hyperref}

\usepackage[capitalize]{cleveref}
\crefname{section}{Sec.}{Secs.}
\Crefname{section}{Section}{Sections}
\Crefname{table}{Table}{Tables}
\crefname{table}{Tab.}{Tabs.}

\def\cvprPaperID{2170} % *** Enter the CVPR Paper ID here
\def\confName{CVPR}
\def\confYear{2022}

\begin{document}

\title{Point Cloud Color Constancy}
% \vspace*{-2ex}
\author{\hspace{7mm} Xiaoyan Xing$^{1,}$\footnotemark[1]~, Yanlin Qian$^{2,}$\footnotemark[1]~, Sibo Feng$^{2,}$, Yuhan Dong$^{1,}$\footnotemark[2]~, and Jiri Matas$^{3}$\hspace{7mm}\\
~~~$^{1}$Tsinghua University, $^{2}$Independent Researcher, $^{3}$Czech Technical University \\
}
% \thanks{
% $^{*}$: Equal contribution\\
% $^{\dag}$: Corresponding author}
\twocolumn[{%
\renewcommand\twocolumn[1][]{#1}%
\maketitle
\begin{center}
\resizebox{\textwidth}{7.3cm}{
\framebox{
\includegraphics[width=\textwidth]{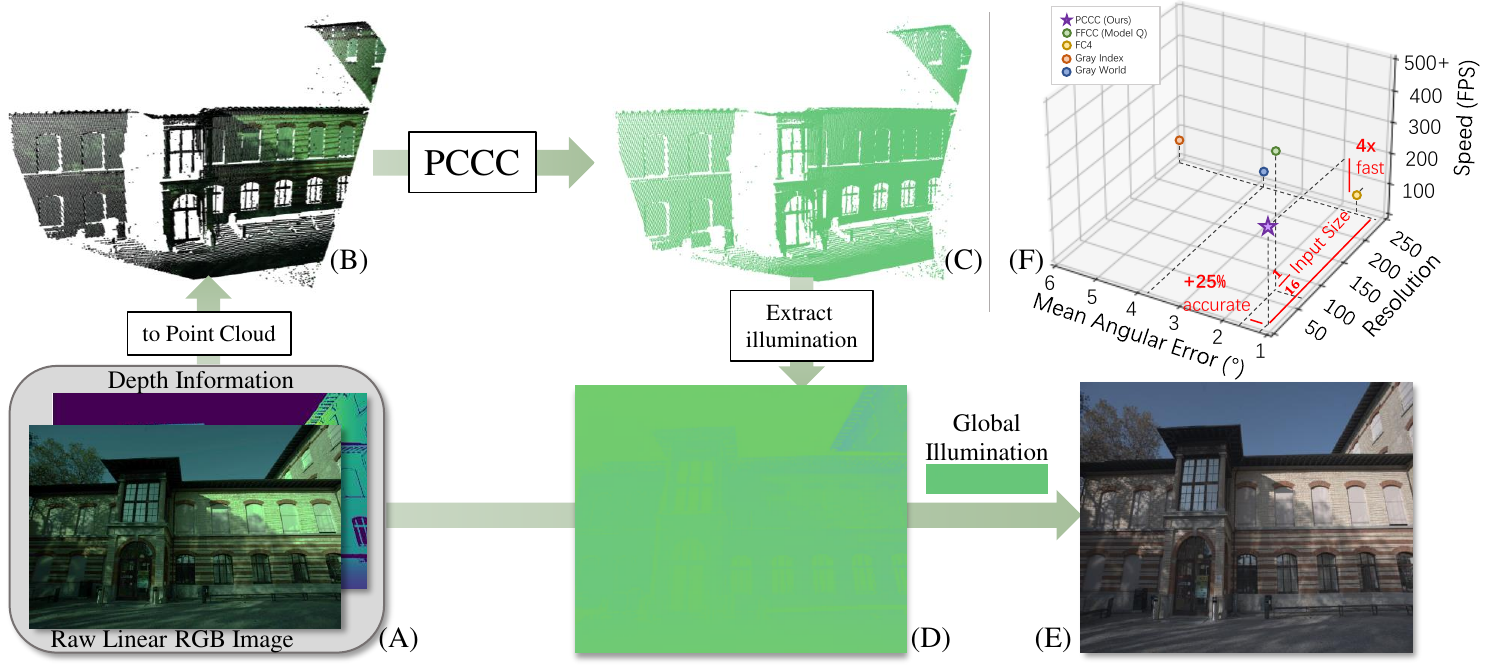}}
}
\vspace{0.3mm}
\captionof{figure}{We present \textit{Point Cloud Color Constancy} (PCCC), which is simple, efficient and hardware friendly. (A): Raw RGB-D image. (B): RGB point cloud generated by (A). (C): Point Cloud with illumination rendered from PCCC. (D) 2D illuminant map extracted from (C). 
(E) Color corrected image. 
(F): Speed, accuracy and parameter number comparison with state-of-the-art color constancy methods.}
\vspace{1mm}
\label{fig:teaser}
\end{center}%
}]
\maketitle
\renewcommand{\thefootnote}{\fnsymbol{footnote}}
\vspace*{-3ex}
\footnotetext{$^*$Equal contribution$/^\dag$Corresponding author. Code \& Data: \url{https://github.com/xyxingx/Point-Cloud-Color-Constancy}} 
% \vspace*{-3ex}
\renewcommand{\thefootnote}{1}

%%%%%%%%% ABSTRACT
\begin{abstract}
\vspace*{-2ex}
   In this paper, we present Point Cloud Color Constancy, in short PCCC, an illumination chromaticity estimation algorithm exploiting a point cloud. We leverage the depth information captured by the time-of-flight (ToF) sensor mounted rigidly with the RGB sensor, and form a 6D cloud where each point contains the coordinates and RGB intensities, noted as (x,y,z,r,g,b). PCCC applies  the PointNet architecture to the color constancy problem,  deriving the illumination vector point-wise and then making a global decision about the global illumination chromaticity. On two popular RGB-D datasets, which we extend with illumination information, as well as  on a novel benchmark, PCCC obtains lower error than the state-of-the-art algorithms.  Our method is simple and fast, requiring merely $16\times 16$-size input and reaching speed over $140$ fps (CPU time), including the cost of building the point cloud and net inference.  
%   Codes \& data: \url{https://github.com/xyxingx/Point-Cloud-Color-Constancy} 
\vspace*{-3ex}
\end{abstract}

%%%%%%%%% BODY TEXT
\section{Introduction}

There are over a billion  smartphone cameras capturing photographs, where the procedure of perceiving true colors of the real world happens in the ISP pipeline. In general, a good-looking picture relies on a single CMOS sensor with a certain color filter array. On flagship smartphones, there is a trend of leveraging two or more digital sensors to ``compute'' a higher-quality image.  Such cases include \{main camera, tele camera\} \cite{abdelhamed2021leveraging}, \{main camera, near inferred sensor\} \cite{zhang2008enhancing}, and \{color camera and time-of-flight (ToF) depth sensor\} \cite{yan2021depthtrack}. In this work,  we proceed in this direction  and use a color camera and a ToF sensor. A ToF depth sensor has been commonly employed for other applications; \eg, auto focus, bokeh effect and 3D face and skeleton scanning.  Here we show a novel task based on combing the RGB and ToF sensors -- RGB-D or point-cloud illumination estimation. 

%what is awb 

Illumination estimation refers to the task of estimating the normalized illumination chromaticity given a raw image, which itself is a core task in computational color constancy.  Color constancy is an intrinsic property of the human eye, which perceives the world independently of illumination color changes.  For technical reasons and engineering benefits, in digital cameras, illumination estimation and a channel-wise rescaling are combined in the so-called auto-white balance (AWB), to realize ``color constancy'' computationally.  An accurate estimation, with proper rescaling factors than render a neutral surface white.   

%why ToF in awb. how other do it. nobody did this. 
%how we do exactly
In common the use-case, illumination estimation only relies on a single raw format image. Here we advocate the use of depth sensor which casts the task into a 3D regression problem, which we show  delivers more accurate results. 
We introduce depth based  color constancy for the following reasons: 1. image statistics are steered by depth information \cite{lu2009color}. More specially, the surface geometry, the texture size and the signal-to-noise ratio varies with depth, which will influence the performance of illumination estimation algorithm; 2. abrupt depth changes correlate with non-smooth illumination distribution. Imagine walking from outside to an indoor room with a tungsten light. In a 2D image, the depth and illumination change dramatically in the area close to the door. 
As shown in Figure \ref{fig:teaser}, the captured RGB-D images are reprojected to form the colored point cloud, on which the proposed method estimates the illumination  for each point. Then these point predictions (Figure \ref{fig:teaser}(C)) are averaged with learnt weights. 
%Considering the accessibility of depth sensor in smartphone, 
%We design a point cloud illumination estimation method as shown in Figure \ref{fig:teaser}.

In a summary, the contributions of the paper are:
\begin{compactitem}
\item We formulate the generic problem of illumination estimation in a 3D world, relaxing the over-simplistic assumptions about uniform illumination for a plain 2D image adopted in prior work
% \footnote{Note that \it{knowing} the illumination is constant is a significant constraint (if the assumption is correct, and exploited, results improve).}
.
  
\item We present PCCC architecture, derived from PointNet,  a novel point cloud net for the color constancy task. We show its superior performance on illumination estimation on three color constancy benchmarks. Side applications, \eg, local AWB is reported as well. 

\item PCCC operates well on the sparse-sampled thumbnail point clouds, which allows over 500 frames per second (include time of building point cloud) on a Nvidia V100 GPU and is hardware friendly for a mobile SoC. 

\item We annotated the illumination groundtruth labels for three popular RGB-D datasets (NYU-v2 dataset~\cite{silberman2012indoor}, Diode dataset~\cite{vasiljevic2019DIODE} and ETH3D dataset\cite{2017ETH3d}) and collected DepthAWB, a novel RGB-D dataset for the color constancy task.

% \textcolor{blue!60}{can we put this paragraph into item2 and demonstrate the third contribution as the sparse point cloud illuminant estimation which is hardware friendly?}

% \item We show that the popular SFU Gray Ball dataset~\cite{ciurea2003large} is suitable for the temporal setting, and introduce
%   a Temporal SFU Gray Ball benchmark. We experimentally evaluate
%   state-of-the-art methods for temporal color constancy on the new benchmark.
  
% On two popular RGB-D datasets with our annotated illumination label and our collected benchmark, PCCC obtains lower error over the state-of-the-art works.
 
\end{compactitem}

\section{Related Work}

The prior art in computational color constancy falls into the learning-based and statistics-based groups. In this section, we consider the candidate method from a different perspective, i.e., the input requirement. 
%how many information flows as input. 

\paragraph{Methods relying on a single Image}

The majority of color constancy methods belong to this category.  Firstly, there exist a list of traditional methods, including Gray World~\cite{buchsbaum1980spatial}, White Patch \cite{brainard1986analysis}, General Gray World \cite{barnard2002comparison}, Gray Edge \cite{van2007edge}, Shades-of-Gray \cite{finlayson2004shades}, LSRS \cite{gao2014eccv}, PCA \cite{cheng2014illuminant} and Grayness Index~\cite{Qian_2019_CVPR}, \etc. They are easy to be implemented in a ISP chipset and may fail if their assumption is violated.  Secondly, more methods are driven by the availability of color constancy datasets which provide carefully annotated illumination groundtruth;   \cite{gijsenij2010generalized,barron2015convolutional, chakrabarti2015color,chakrabarti2012color,finlayson2013corrected,gehler2008bayesian,gijsenij2011color,hu2017cvpr,joze2014exemplar,shi2016eccv} learn a high-dimensional mapping from the capture raw data to the sought illumination vector after optimization on training set. The learning tools behind includes gamut mapping, least square minimization, random forest,  neural networks and so on.

% \cite{chakrabarti2012color,gijsenij2010generalized,gehler2008bayesian,gijsenij2011color,joze2014exemplar,yanlin2016icpr,yanlin2017iccv} aim at building a model that relates the captured image $I$ and the sought illumination $L$ from extensive training data. Among the best-performing state-of-the-art approaches, the CCC method \cite{barron2015convolutional} discriminatively learns convolutional filters in a 2D log-chroma space. This framework was subsequently accelerated using the Fast Fourier Transform on a chroma torus~\cite{barron2017fourier}. Chakrabarti \textit{et al.}~\cite{chakrabarti2015color} leverage the normalized luminance for illumination prediction by learning a conditional chroma distribution.
% DS-Net~\cite{shi2016eccv} and FC$^4$ Net~\cite{hu2017cvpr} are two deep learning methods, where the former chooses an estimate from multiple illumination guesses using a two-branch CNN architecture and the later addresses local estimation ambiguities of patches using a segmentation-like framework. 
% Learning-based methods achieve great success in predicting pre-recorded ``ground-truth'' illumination color fairly accurately, but heavily depending on the same cameras and/or scenes being in both training and test images (see Sec. \ref{sec:novelgi} and Sec. \ref{sec:agnostic}). The Corrected-Moment method \cite{finlayson2013corrected} can also be considered as a learning-based method as it needs to train a corrected matrix for each dataset. 

\paragraph{Methods relying on Image(s) and Auxiliary Information} 
There is a research trend of including more information else than a single image in design of color constancy methods, for special purposes. Abdelhamed~\etal~\cite{abdelhamed2021leveraging} leveraged main camera and tele camera simultaneously in a very light-weight neural network, releasing the discriminative ability of color filter array correlation. Barron~\etal~\cite{barron2017fourier} proposed a Fourier transform-based model (model O in the paper), learning the model weight from the camera metadata like aperture and so on.  Qian~\etal~\cite{yanlin2017iccv,qian2020benchmark} claimed the preceding image sequence benefits the illumination estimation for the shot image. Yoo~\etal~\cite{yoo2019dichromatic} explored the AC light source in a high-speed setting. Ebner~\etal~\cite{ebner2014improving} firstly estimated the illumination chromaticity by combing Gray World with depth map.

Differing from \cite{ebner2014improving}, in a end-to-end way, our approach learns how to use depth for referring illumination  from a sparsely-sampled point cloud. As shown in Table~\ref{tab:advantage}, it also supports semantic info extraction and 3D point-wise AWB.

\section{Point Cloud Color Constancy}
\begin{figure*}
    \centering
    \includegraphics[width=\linewidth]{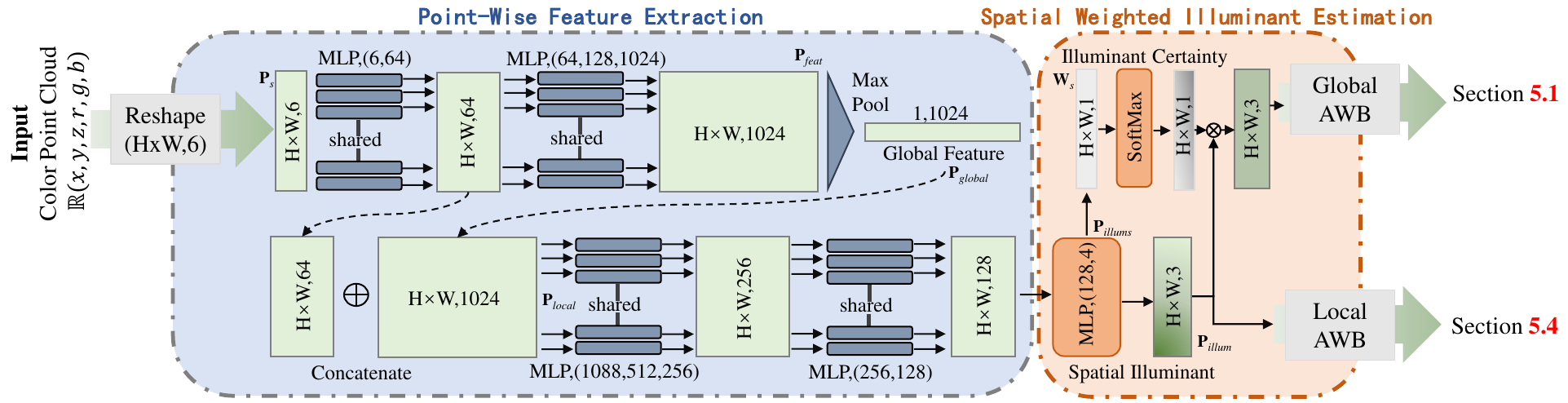}
    \caption{The PCCC architecture consists of a point-wise feature extraction block - a slight modification of  PointNet\cite{qi2017pointnet}, and  the spatial weighted illuminant estimation block. Given $H \times W$ points as input, PCCC outputs a spatial weighted illuminant, which benefits global auto white balance, AWB (Section \ref{Sec:QR}). With a slight setting change, we can also achieve point-wise illumination estimation (Section \ref{Sec:Local}).}
    \label{fig:pipline}
\end{figure*}
%In this section, we introduce the various steps of our point-cloud illumination estimation algorithm: 
We describe PCCC in the following order:
1) reformulating the illuminant estimation problem on RGB-D images (Section \ref{Sec:problem}); 2)  aligning the RGB images and depth maps spatially (Section \ref{Sec:alig}); 3) generating point cloud with tristimulus color values (Section \ref{Sec:pcd}); 4) explaining our point-cloud illuminant estimation network (Section \ref{Sec:Regressor}); and 5) proposing two augmentation tricks based on vibrating camera position and illumination chromaticity (Section \ref{Sec:Aug}).
%a camera position-based augmentation and a spatial illuminant augmentation (Section \ref{Sec:Aug}).
\subsection{Formulation}
\label{Sec:problem}

Given an RGB image $\textbf{I}$, the image formation can be described as follow:
% \begin{equation}
%     \textcolor{black}{\textbf{I} = \int_{\omega_i \in \Omega+} \int_{\lambda} (N, \omega_i) E(\omega_i,\lambda)S(\omega_i,\omega_o,\lambda)R_{RGB}(\lambda)d\lambda d\omega},
% \end{equation}
% where $\omega_i$ is the lighting angle from the upper hemisphere $\Omega+$, $\omega_o$ is the viewing angle and $N$ is the surface normal. $E$ refers to illuminant spectral power distribution, $S$ is the surface reflectance function, and $R_{RGB}$ refers to the sensor responses. Our aim in this paper is to get illuminant $\textbf{E}=\int_{\omega_i \in \Omega+} \int_{\lambda}  E(\omega_i,\lambda)R_{RGB}(\lambda)d\lambda d\omega$ for the captured environment, or even more ideally, for each point position. 
\begin{equation}
    \textcolor{black}{\textbf{I} = \int_{\lambda}  E(\lambda)S(\lambda)R_{RGB}(\lambda)d\lambda},
\end{equation}
where $\lambda$ is the visible wavelength.
$E$ refers to illuminant spectral power distribution, $S$ is the surface reflectance function, and $R_{RGB}$ refers to the sensor responses. Our aim in this paper is to get illuminant $\textbf{E}= \int_{\lambda}  E(\lambda)R_{RGB}(\lambda)d\lambda$ for the captured environment, or even more ideally, for each point position. 

% If we extend the space of intergral from $\Omega\in\mathbb{R}^{U\times{V}\times3}$ to $\Omega_{pcd}\in\mathbb{R}^{X\times{Y}\times{Z}\times{3}}$, we obtain 3D-version lambertion model.
%a spatial imaging formation of a 3-D space or of a point cloud.  

% Similar to the conventional computational color constancy methods. Our goal is to estimate the global illuminant $E_{global}$ from inputs, the difference is our inputs are point clouds.
% Since the viewing angle and surface normal are hard to extract through a single 2D image, introducing depth information to color constancy task may help. 
Based on prior works~\cite{ebner2014improving,lu2009color} and our observation, with point clouds (or images with depth information) as inputs, the color constancy method benefits from the following: 1) explicitly discriminate the ambiguity illuminant through depth discontinuity; 2) identify the dominant illuminant under different depth; 3) simulate different viewing angle of the same scenes (Section \ref{Sec:Aug} and \ref{Sec:Ablation}) ; and 4) estimate global illuminant based on spatial illuminant distribution (Section \ref{Sec:Regressor}, \ref{Sec:Ablation} and \ref{Sec:Local}).

\subsection{Alignment between RGB and Depth Map}
\label{Sec:alig}
RGB images and depth maps are captured by different sensors separately, thus alignment is needed. 
% Our alignment of RGB and depth images can be divided into the following steps: 
The process of alignment follows as:
1) calibration two cameras' intrinsics and extrisics using Zhang~\etal~\cite{zhang2000flexible}; 2) calculate the rigid transformation from coordination of RGB camera to the coordination of depth camera, and 3) 
colorize the depth map using the corresponding RGB intensities.
Since the rigid connection of two sensors holds when mounted, the transformation matrix is pre-computed before dataset collection. 

\subsection{Point Cloud with Tristimulus Values}
\label{Sec:pcd}
Given an RGB image $\textbf{I} = [\textbf{I}_r,\textbf{I}_g,\textbf{I}_b] \in\mathbb{R}^{U\times{V}\times3}$ and its corresponding depth map $\mathrm{D}\in\mathbb{R}^{U\times{V}\times1}$, we aim to build  a colored point cloud denoted as $\textbf{P} = \{\textbf{p}_i | i \in 1,,,n\}$.  Each point $\textbf{p}_i$, as a vector of (x,y,z,r,g,b) values, is computed as:
\begin{equation}
    \textbf{p}_i = (\frac{(u-c_x)d}{f_x}, \frac{(v-c_y)d}{f_y}, d,r,g,b),
\end{equation}
where {u,v} are the 2D coordinates in $\textbf{I}$ for the $i_\text{th}$ pixel, $d$ is the depth value, $\{f_x$,$f_y\}$ represent the focal lengths, ($c_x$, $c_y$) the principal point. $\{r,g,b\}$ refer to the color channels in $\textbf{I}$.

% a colored point cloud $\textbf{P}_c = [\textbf{P}_r,\textbf{P}_g,\textbf{P}_b]\in\mathbb{R}^{X\times{Y}\times{Z}\times{3}}$ in the camera coordinate system can be transformed by following formula:
% \begin{equation}
%     \mathrm{Z}\textbf{I}_{r}=\left[\begin{array}{ccc}
%     f_x&0     &c_x  \\
%     0&  f_y   &c_y \\
%     0&  0   &1
%     \end{array}\right]\textbf{P}_{r}
% \end{equation}
% where $[f_x,f_y,c_x,c_y]$ are the intrinsics, $[r,g,b]$ indicates each color channel respectively. 

Comparing to 2D images, the colored point clouds carry more rich information of inputs such as spatial distance between objects, spatial color transformation, and explicit geometric structure. 
% Moreover, several prior researches \cite{danelljan2016probabilistic} have shown color information is important yet efficient for point cloud operation. Which inspire us to see the possibility of estimating the illuminant from point clouds.\textcolor{red}{this sentence should be in related work. } 
Unless explicitly stated otherwise,  ``point cloud'' in this paper denotes colored point cloud. 
% \subsection{Spacial Illuminant Estimation Network}
\subsection{Illumination Regressor on Point Cloud}
\label{Sec:Regressor}
Given a dataset of $N$ point clouds, $\mathcal{P}=\{\textbf{P}_{1},\textbf{P}_{2},...,\textbf{P}_{M}\}$ and the corresponding  global illumination label $\mathcal{E}=\{\textbf{E}_1,\textbf{E}_2,...,\textbf{E}_M\}$, we then can train a neural network Given a dataset of $N$ point clouds, $f_{\theta}:\mathcal{P}\to{\mathcal{E}}$
\begin{equation}
\label{Equ:NN}
    \hat{\textbf{E}_i} = f_{\theta}(\textbf{P}_i),
\end{equation}
where $\hat{\textbf{E}}$ is the predicted illumination vector.

To estimate the dominant illuminant $\textbf{E}$, we 1) extract the lighting feature out of the point set; 2) obtain the approximately illuminant distribution, and 3) estimate the dominant illuminant from the spacial distribution.

\textbf{Feature extraction} 
Point clouds are spatial sparse and disorderly. Traditional CNN architectures are designed for 2D images (which are in regular format) and may not fit the task. We employ PointNet \cite{qi2017pointnet} as our feature extraction module, which is designed for the point cloud-based classification task. PointNet subtly overcomes the unordered issue with an architecture that consists of several multi-layer perceptrons (MLP) and max pooling as its basic layers. 

To feed MLP, $\textbf{P}$ is converted to an unordered sequential of $N$ points,  $\textbf{P}_{s}\in\mathbb{R}^{N\times6}$, where 6 channels represent the camera spatial coordinate $\{X,Y,Z\}$ and corresponding color values $\{R,G,B\}$ for each point. Firstly, MLP transforms the $\textbf{P}_{s}$ of 6 channels into $\textbf{P}_{feat}$ of 1024 channels, while $N$ remains the same. Then, max pooling aggregates $\textbf{P}_{feat}$ with $N$ points into one single global feature $\textbf{P}_{global}$. 

\textbf{Spatial weighted illuminant estimation}
Technically, the MLP can decrease the channel of global feature $\textbf{P}_{global}$ from 1024 to 3, yielding the final prediction, which is simple but loses local estimates.
%and achieve the illuminant estimation by suitable loss function. Which is simple but loses the local information. 
Inspired by \cite{hu2017cvpr}, we devise a spatially weighted estimation module, by introducing the point-cloud weight matrix $\textbf{W}_s$ in stage of network inferring.

% In the original PointNet, max pooling is employed to aggregate $\textbf{P}_{feat}$ to one point $\textbf{P}_{max}\in\mathbb{R}^{1\times1024}$, which is simple but miss the local information. To overcome, we design a spatially weighted illuminant estimation module, by implicitly introducing the 3-D weight matrix $\textbf{W}_s$.

% PointNet offers a pixel-wise setting by repeating the global feature $\textbf{P}_{global}$ and concatenating it to the intermediate output from first MLP group. 
% We adopt the strategy to extract the feature point-wisely . 

To include more local information, feature extraction is boosted by aggregating the global feature $\textbf{P}_{global}$ and the intermediate output from first MLP group to be  $\textbf{P}_{local}\in\mathbb{R}^{N\times{1088}}$. 
Then $\textbf{P}_{local}$ is reduced to $\textbf{P}_{illums}=[\textbf{P}_{illum},\textbf{W}_{s}]\in\mathbb{R}^{N\times{4}}$ by MLP, where $\textbf{P}_{illum}\in\mathbb{R}^{N\times{3}}$ represents the illumination estimation of each point, and spatial illuminant matrix $\textbf{W}_s\in\mathbb{R}^{N\times{1}}$.
% and the forth channel  is the spatial illuminant matrix $\textbf{W}_s\in\mathbb{R}^{N\times{1}}$. 
Mathematically global illuminant $\hat{\textbf{E}}_{global}\in\mathbb{R}^{1\times3}$ can be achieved by : 
\begin{equation}
    \hat{\textbf{E}}_{global} = \sum_{R,G,B}(\mathrm{softmax}(\textbf{W}_s)\cdot\textbf{P}_{illum}),
\end{equation}
where softmax normalizes $\textbf{W}_s$ into \{0,1\} as a probability mask. The global illuminant is employed for global color constancy. The probability mask and local illuminant can later be leveraged for the trial in local AWB.
% FC4 \cite{hu2017cvpr} proved mathematically the implicit weight map can prevent the network from learning noisy data in 2D illuminant estimation. Inspired by FC4, we utilize MLP to realize pointwise weighting in point cloud. Lower error is obtained since included (Section \ref{Sec:Ablation} ). 

% We revisit the implicit weight structure in point cloud, and find it improve the illuminant estimation in pointclouds and illustrate the three-dimensional illumination distribution (Section \ref{Sec:Ablation} and \ref{Sec:Local}). 

\textbf{Loss Function}
Similar to the prior arts, we use the recovery angular loss $\mathcal{L}_{ang}$ as our loss function: \begin{equation}
\label{Equ:ang}
    \mathcal{L}_{ang}(\hat{\textbf{E}}
    ,\textbf{E}) = \arccos({\frac{\hat{\textbf{E}}\cdot{\textbf{E}}}{||\hat{\textbf{E}} ||\cdot{||\textbf{E}||}}}),
\end{equation}
where $\hat{\textbf{E}}$ is the predicted illumination vector and $\textbf{E}$ is the ground truth illumination label.  With the loss, our net converges in an end-to-end way.

\textbf{Discussion}
Theoretically, any network that design for point clouds \cite{li2018pointcnn,qi2017pointnet++} or for RGB-D input can be employed to replace the PointNet-based net here. However, due to the nature of the color constancy problem, we argue that an ideal point cloud color constancy network should 1) have a simple but effective structure that keeps good trade-off between accuracy and running speed; 
%obtains top accuracy in real-time speed; 
2) be not sensitive to the noise and sparse sampling rate and
%illuminant difference, which can handle different illuminant under same scenario and 
3) in default working with thumbnail-size input, which is usually provide by specialized ISP hardware node.
Quantitative experiments to the mentioned properties of our design are in Section \ref{Sec:Why}.

\subsection{Point Cloud Augmentation}
\label{Sec:Aug}

\begin{figure}[t]
    \centering
    \includegraphics[width=\linewidth]{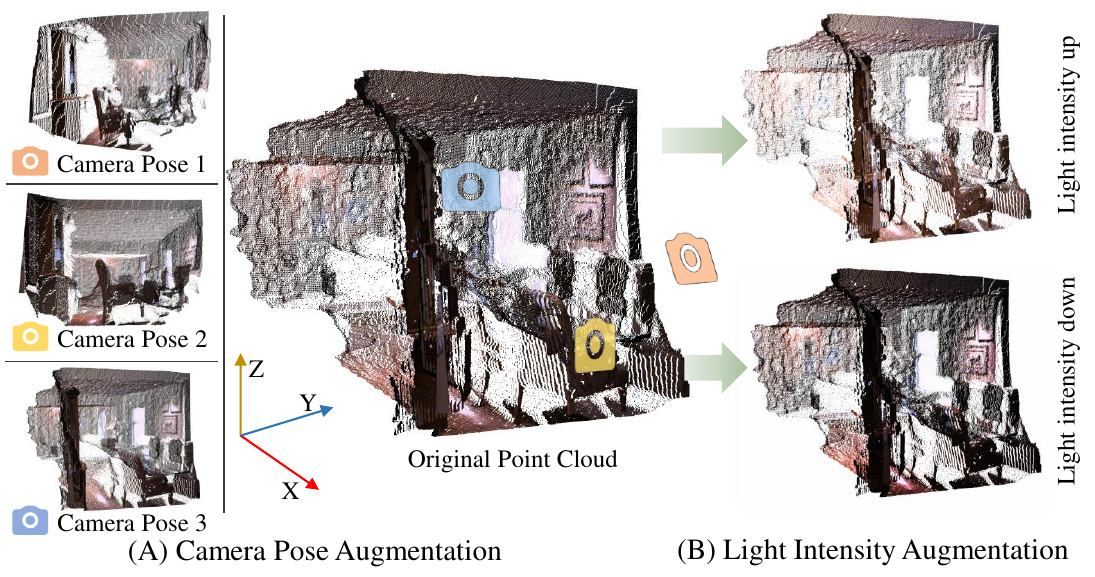}
    \caption{We use two augmentation approaches: (A) camera pose augmentation simulates images capture from a different viewing angle. (B) light intensity augmentation generates new pixel-wise illumination field under the same illuminant.}
    
    \label{fig:augmentation}
\end{figure}
%Point clouds give the opportunity to augment data on a spatial level. 
Point cloud allows us to augment data from a 3D prospective. We propose two augmentation methods based on our observation that illumination on a point is roughly invariant to mild \textit{camera pose} jittering and \textit{light intensity} changes. These methods rely on separating a point cloud into two parts 1) coordination value $\textbf{P}_{xyz}$ and 2) color value $\textbf{P}_{RGB}$, considering the order of point cloud is fixed during our augmentation period. We can augment the camera position and light intensity separately as following. 

% we propose a camera-based spacial augmentation to increase the number of training samples and to avoid over-fitting during training. 
% Our augmentation is based on a series of observations of real photo taking. We assume that 1) the dominant illuminant in a scene does not have a major shift; 2) the objects in a scene do not change much. 

\textbf{Camera Pose Augmentation}
 is inspired by that small rotation does not affect the decision of the dominant illumination color. 
 As shown in Figure~\ref{fig:augmentation}, we sample a rotation matrix $\textbf{R}\in{SO(3)}$ to simulate the scenarios where we take images from different perspectives. 

% \textbf{Camera Pose Augmentation}
%  is inspired by the fact that rotation does not affect the dominant illumination chromaticity which lies in a continuous space, we introduce a rotation matrix $\textbf{R}\in{SO(3)}$ to simulate the different perspectives of image capturing. 
 
\textbf{Light Intensity Augmentation} is inspired by luminance can be changed when the relative distance of the illuminated objects and the light source changes. 
Based on this, we use an intensity scaling factor $L$ to realize luminance-based augmentation. 
%Considering this, we propose light intensity factor $L$ to achieve a simple yet efficient augmentation on Illuminance. 
% Given a 6-D point cloud $\textbf{P}_c=[\textbf{P}_{pos},\textbf{P}_{color}]$, we introduce a rotation matrix $\textbf{R}\in\mathbb{R}^{3\times3}$ to simulate the different angle of image shooting. 

% The point cloud $\textbf{P}_c$ is a series of points in euclidean space. Here we introduce $\theta_x,\theta_y$, and $\theta_z$ to represent the rotation angle in Pitch, Yaw, and Roll axis, respectively. 
% We use matrix $\textbf{R}$ to simulate the photo taken at same scene but with different angles of view. 

Considering both augmentation ways, the new point cloud $\textbf{P}^{'}$ is:

%Finally, the new point cloud $\textbf{P}^{'}$ under the two augmentations can be formulated as:
% \begin{equation}
%     \textbf{R}_{x,y,z} = \left[\begin{array}{ccc}
%      1&0  &0  \\
%     0&\cos{\theta_x} & -\sin{\theta_x}\\
%     0& \sin{\theta_x}&\cos{\theta_x}
%     \end{array}\right]\left[\begin{array}{ccc}
%      \cos{\theta_y}&0  &\sin{\theta_x}  \\
%     0& 1 & 0\\
%   -\sin{\theta_x}&0&\cos{\theta_y}
%     \end{array}\right]
% \end{equation}
% \begin{equation}
%     \textbf{R}_{x,y,z} = \textbf{R}_{x}(\theta_x)\textbf{R}_{y}(\theta_y)\textbf{R}_{z}(\theta_z),
% \end{equation}
% then the rotated point cloud $\textbf{P}_{c|\mathrm{R}}$ can be obtained by
\begin{equation}
    \textbf{P}^{’}=\textbf{R}\textbf{P}_{xyz}\oplus{L\textbf{P}_{RGB}},
\end{equation}
where $\textbf{R}$ is the sampled rotation matrix, $L$ is a scalar from a normal distribution $\mathcal{N}(1,0.15^2)$, and $\oplus$ is channel-wise concatenation. 
% It should be noticed that, the angle of rotation obey a normal distribution and is limited by the effective area of view. 

% \textbf{Light Intensity Augmentation.} Illuminance can be changed when objects get closer or further to the light source, based on that, we propose a simple yet efficient augmentation on light intensity. Given a 6-D point cloud $\textbf{P}_{c}$ we can increase or reduce the light intensity on each point through a global as $\textbf{P}_{c|L} = \textbf{P}_{c}L$, where $L$ obeys an 1-D normal distribution, and $\sigma=0.15, \mu=1$.

% Plentiful yet accurate augmentation in illuminant could improve the network learning ability. Inspired by \cite{abdelhamed2021leveraging,lo2021clcc}, we further design an illuminantion augmentation in 6-D point clouds. Given a raw-RGB Image $\textbf{I}_{RGB_1}$ under an illuminant, $\textbf{E}_1$, we can calculate the 24 color patches through the color checker board, $C_1\in\mathbb{R}^{24\times3}$. Then, we can calculate a color transformation matrix, $T_{C_{1\to{2}}}\in\mathbb{R}^{3\times3}$, between another image $\textbf{I}_{RGB_2}$ under different illuminant $\textbf{E}_2$ from the same camera. Finally, we can obtain a new image $\textbf{I}_{RGB_1^{k}}$ under different illuminant $\textbf{E}_k$ by,
% \begin{equation}
%     \textbf{I}_{RGB_1^{k}} = \textbf{I}_{RGB_1}T_{C_{1\to{k}}}.
% \end{equation}

% Given $N$ point clouds, with the illumination augmentation above, we can obtain $N^2-1$ point clouds with different illuminations.

\section{Dataset Preparation}
\label{Sec: datas}
To train PCCC, the pairs of the point sets (from RGB-D images) and the corresponding illumination labels are needed. 
However, to our knowledge, there are no such datasets publicly available. 
To evaluate our method, we first present in Section \ref{Sec:NYU} reannotating the commonly used depth datasets NYU-v2~\cite{silberman2012indoor} and Diode~\cite{vasiljevic2019DIODE}. 
Then in Section \ref{Sec:ETH} we relabel the ETH3D dataset~\cite{2017ETH3d}, which includes sensor raw files. 
Finally and the most importantly, we show the collection of our point cloud illumination dataset using a single lens digital camera and an accompanied  Intel laser depth sensor. 
% we describe how we collect our own dataset with both RGB-D images and the ground thruh from the color checker, in Section \ref{Sec:Depth-awb}.  

\subsection{Annotating NYU-v2 \& Diode Depth Dataset}
\label{Sec:NYU}
NYU-v2 \cite{silberman2012indoor} is a classical RGB-D dataset, with hundreds of indoor scenes captured by Microsoft Kinect. Diode \cite{vasiljevic2019DIODE} is a latest RGB-D dataset released by Toyota Technological Institute at Chicago, this dataset contains more than 20 scenes include indoor and outdoor scenes. 
Affected by the illuminant and position, there are numbers of AWB-biased images in these datasets (See supplement material). % % (a show of incorrect awb images) 
We found these images are suitable for us to label the correct illumination according to the neutral surfaces and validate the color constancy algorithm.

Images from these two datasets are in sRGB. To label the ground truth illuminant, we process images by the following steps: 1) switch sRGB to LRGB using inverse gamma operation~\cite{anderson1996proposal,ebner2007color}; 2) remove the tuning operation using a $3\times3$ matrix as suggested in \cite{afifi2019color}; and 3) label out the neutral surface area (which is deemed as neutral without doubt, like white printed paper) and calculate the illumination vector (Details in supplement materials). 
The steps can be formulated as:
\begin{equation}
    \textbf{E}_{label} = \Gamma{(\textbf{M}^{-1}\times{\textbf{I}_{sRGB}^{1/\gamma}})},
\end{equation}
where $\Gamma(\cdot)$ represents the area selection and illuminant calculation. $\textbf{E}_{label}$ is the labeled ground truth illumination, $\textbf{M}$ is the tuning matrix of images.

After the preprocessing steps, we obtain linearization images with labels from two datasets. These images contain indoor and outdoor scenes, with diverse label distribution.  

\textbf{Discussion}  We acknowledge that relabeled illumination on sRGB images may not be as accurate as that in raw LRGB images, and our transformation can not really transform the sRGB to LRGB, which is a rough operation and may bring nosie to the images. Since all color constancy methods will surfer the same bias on the datasets, we argue that these data are of experimental value for us to test our method (See Section \ref{Sec:Experiments} for more experimental details).  

\subsection{Annotating ETH3D Depth Dataset}
\label{Sec:ETH}
ETH3D dataset~\cite{2017ETH3d} is a stereo RGB-D dataset with 13 different scenes with indoor and outdoor, the RGB images are captured by a Nikon D3 camera, and the depth maps are from a radar. 
% There are also some incorrect white balance setting images (See Figure xx). 
The dataset provides the raw format of images. We can process the images using LibRaw, and obtain the raw linear RGB images, which makes them closer to the images we used in camera auto white balance pipeline. 

We use the similar label method as Section \ref{Sec:NYU} to obtain the illuminant. Considering we already have raw DSLR images $\textbf{I}_{Raw}$, the labeled illuminant $\textbf{E}_{label}$ can be obtained by:
\begin{equation}
      \textbf{E}_{label} = \Gamma{(\textbf{I}_{Raw})}.
\end{equation}

% The raw images, ground truth illuminant and depth map, are obtained after the processing step.  

\subsection{DepthAWB -- Real Color Constancy Dataset with Depth}
\label{Sec:Depth-awb}

\begin{figure}[t]
    \centering
    \includegraphics[width=\linewidth]{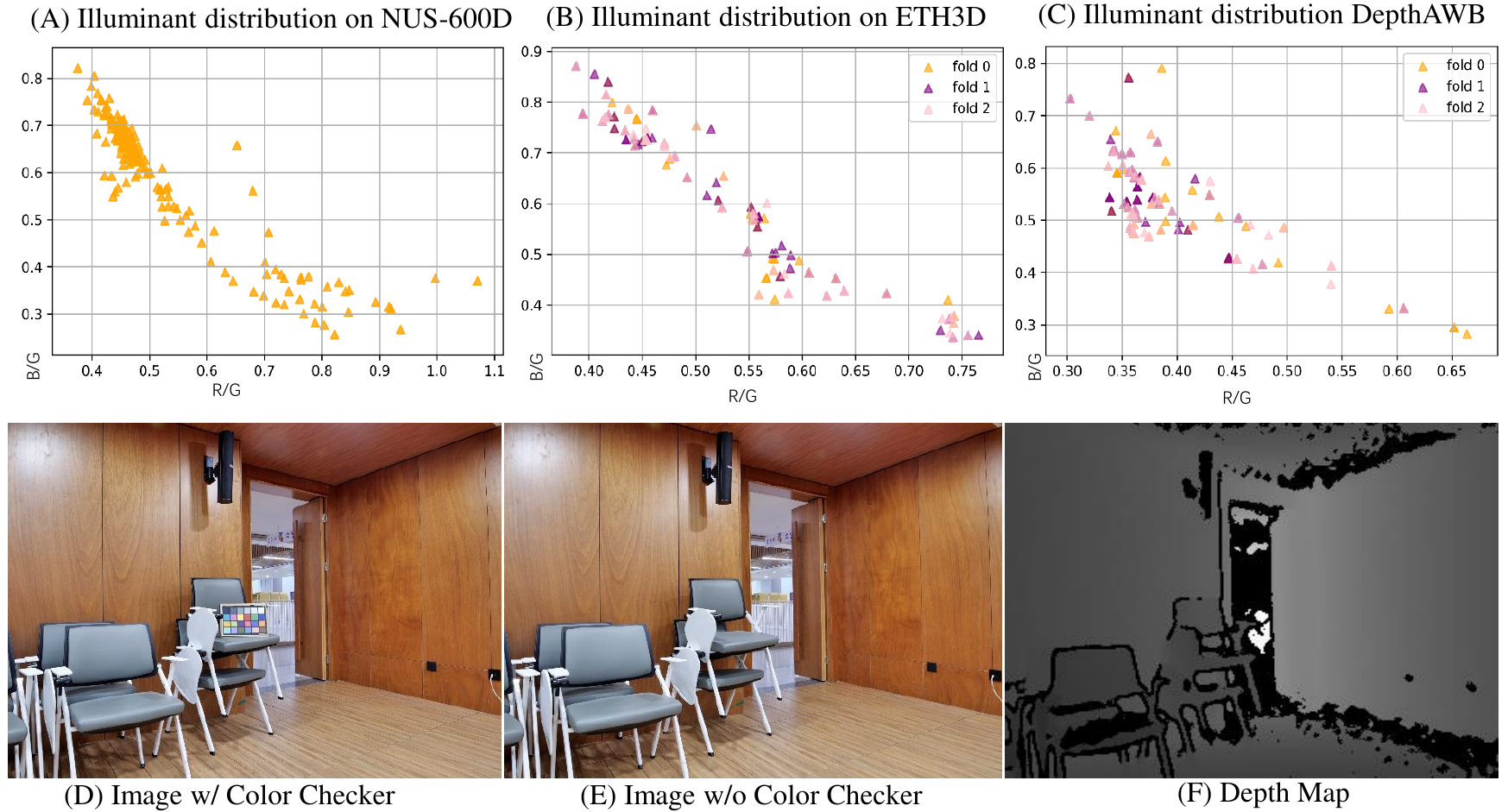}
    \caption{{(A)-(C): Illumination distribution of NUS-600D dataset, our relabel ETH3D dataset and propose dataset. (D), (E) a pair of images we captured sequentially. We use (D) with color checker for labelling and (E) for training and testing. (F) ToF depth map.}}
    \label{fig:hardware}
\end{figure}

Despite we have the annotated dataset on raw images, still the ground truth illuminants have biases, since we can not really know the accurate color. Therefore, an RGB-D dataset with raw images and color checker labeled illuminants is needed. Following prior works on RGB-D datasets collection \cite{silberman2012indoor} and color constancy dataset collection \cite{gehler2008bayesian}, we collected our own depthAWB dataset, which to our knowledge is the first RGB-D dataset with labeled illuminant. Our dataset preparation steps are as following:
% 1) RGB images and corresponding depth map collection; 2) RGB camera and depth map calibration; and 3) illuminant annotation.

\textbf{Data collection:} Since RGB-D camera like Kinect-V2 does not provide the raw images, therefore, following \cite{2017ETH3d}, we use LUMIX GH5S camera
\footnote{We fixed focal length and set camera aperture as ${f}/8.0$, while extend the shutter speed, details discussed in supplement.} 
to capture the DSLR RGB images and Intel Resense L515 to capture the depth maps. Our data capture setup is presented in supplement material.

% Due to the limitation of Realsense camera, we mainly capture the indoor scenario, since the indoor scenario often carries more complicated illuminants.  

\textbf{Data calibration:} Calibration is a fundamental operation, as introduced in Section \ref{Sec:alig}, we first capture several images from RGB camera and depth camera, by tilting the a checker board with checker size 30 millimeter vertically and horizontally; then we calculate the transformation matrix using the Caltech camera calibration toolbox \cite{Jean-YvesBouguet}. 

\textbf{Illumint annotation:} Since we obtain the raw images in data collection,  we demosaic the raw images similar as Section \ref{Sec:ETH}, and we directly annotated the ground truth on the xrite color checkerboard.

\subsection{Illumination distribution}
In Figure \ref{fig:hardware} (A)-(C), we compare the illumination distribution of our labeled ETH3D dataset, our collected Depth AWB dataset to the commonly used NUS dataset\cite{cheng2014illuminant} Cannon 600D subset (NUS-600D). Our datasets cover a wide range of illuminant values, and share the same trend in illuminant distribution as NUS-600D.

% Table generated by Excel2LaTeX from sheet 'Sheet3'
\begin{table*}[t]
  \centering
  \resizebox{\textwidth}{!}{
  \begin{threeparttable} 
    \begin{tabular}{lccccc|ccccc|ccccc||rr}
    \toprule
    {Method} & \multicolumn{5}{c|}{Angular Error (NYUv2\&Diode)} & \multicolumn{5}{c|}{Angular Error (ETH3d RAW)} & \multicolumn{5}{c||}{Angular Error (DepthAWB)} & \multicolumn{1}{c}{Training} & \multicolumn{1}{c}{Test} \\
          & Mean  & Med. & Tri. & B25\% & W25\% & Mean  & Med. & Tri. & B25\% & W25\% & Mean  & Med. & Tri. & B25\%& W25\% & \multicolumn{1}{c}{ time (s)\tnote{1}} & \multicolumn{1}{c}{ time (ms) } \\
    \midrule
    % ~~~~~~~~\textit{Learning-free Methods}\\
   GrayWorld~\cite{buchsbaum1980spatial} & 5.39  & 3.83  & 4.37  & 1.23& 11.81  & 3.18  & 2.87  & 2.95  &1.00 & 5.70  & 6.03  & 5.23  & 5.40  & 3.48& 9.75  &     -  & {10.00} \\
    WhitePatch~\cite{brainard1986analysis} & 6.37  & 5.53  & 5.59  &3.77 & 10.46  & 6.28  & 6.41  & 6.49  & 1.20& 13.09  & 11.77  & 12.51  & 12.23  & 3.74& 18.80  &    -   & {7.83} \\
    $1^{st}$GrayEdge~\cite{van2007edge} &   4.02    &2.71     &2.96 & 0.82      &9.43       & 7.81&  5.44     & 6.49      & 1.06      & 17.09      &6.56       &6.13       &6.06 &1.52       &12.83       &    -   & 34.00 \\
   $2^{nd}$GrayEdge~\cite{van2007edge} &  4.15      & 2.90&   3.10    & 0.78      &  9.81     &       8.02&5.30       &6.38       &1.21       &18.20 &   7.12    & 6.91    & 6.97  &    1.87   &    13.07   &    -   &42.50  \\
    Shade of Gray~\cite{finlayson2004shades} & 3.69  & 2.68  & 2.90  & 0.61& 8.59  & 7.25  & 5.02  & 6.23  & 0.95& 15.59  & 5.84  & 4.59  & 5.11  &1.22& 11.82  &    -   & {320.00} \\
    % Gamut Pixel~\cite{2008} &       &       &       & &       &       &       &  &     &       &       &       &      &  &       &       &  \\
    APAP (GW)~\cite{afifi2019projective}  & 3.47  & 2.31  & 2.61  &0.67 & 7.78  & 2.71  & 2.53  & 2.54  &0.82 & 4.69  & 3.69  & 2.69  & 3.08  & 1.70& 6.50  & \cellcolor{pink!60}{16} & {16.00} \\
    % APAP~(SoG)~\cite{afifi2019projective} &       &       &       & &       &       &       &  &     &       &       &       &      &  &       &       &  \\
    Shen et al.~\cite{shen2009simple} & 5.55  & 3.61  & 3.88 & 1.03& 13.39  & 3.90  & 3.12  & 3.31  & 1.08& 7.54  & 5.05  & 3.19  & 3.80 & 0.77 & 11.85  & -      & {37.00} \\
    GI~\cite{Qian_2019_CVPR}    & 4.28  & 3.01  & 3.27& 0.85 & 9.81  & 3.26  & 2.02  & 2.19 &0.49  & 7.35  & 3.91  & 2.09  & 2.58 & 0.49 & 10.23  &  -     & {109.00} \\
    % ~~~~~~~~\textit{Learning-based Methods, Our rerun}\\
    Bayesian~\cite{gehler2008bayesian} &  9.64  & 10.01 &9.59
    &3.61  &15.68    & 7.32       &5.31       &6.02      &1.70  &15.42       &  5.97     & 5.09     &5.24       &1.91       &11.20       &\cellcolor{yellow!60} 35      &2134.42 \\
    Cheng et al.~\cite{cheng2015effective}  &  3.11  & 2.02 &2.27
    &0.59  &7.39    & 3.44       &1.71       &2.06      &0.44  &9.43       & 1.89     & 1.15     &1.24       &0.22       &4.71       & 63      &331.20 \\
    Quasi U CC \cite{bianco2019quasi} &  3.76  & 2.20 &2.54     &0.45  &9.28    & 3.90       &3.29       &3.43       &0.85  &7.58       &  4.45     & 3.80      &3.90       &1.04       &9.03       & -      & 345.13 \\
    FC4~\cite{hu2017cvpr}   & 2.49  &1.61  &1.79  &0.47 & 6.03  & 1.19  &0.90  &0.98  & 0.32 &2.46  & 1.34  & 0.98  & 1.04 &0.24 & 3.08  & {13000} & {38.21} \\
    FFCC (Model J)~\cite{barron2017fourier}  & 2.82  & 1.87  & 2.00 &0.51  & 6.82  & 1.16  & 0.91  & 0.99 & 0.31 & 2.43  & 1.32  & 0.76  & 0.88 & 0.27& 3.21  &  67    & 29.00  \\
    FFCC (Model Q)~\cite{barron2017fourier}  & 2.75  & 1.90  & 2.01 &0.48  & 6.63  &  1.08  & 0.88  & 0.91 & 0.26 &  2.26  & 1.40  & 0.81  & 0.93 & 0.29& 3.40  &  51   &\cellcolor{pink!60}1.10  \\
    % CLCC~\cite{lo2021clcc}  & -  & -  & - &-  & -  & 1.16  & 0.91  & 0.99 & 0.31 & 2.43  & 1.40  & 0.81  & 0.93 & 0.29& 3.40  &       &  \\
        \midrule

    Ours (16 Points, w/o depth)&2.96 & 1.82 & 2.12 &0.63 & 6.91  & 1.58 &0.98& 1.09& 0.22& 3.82  &2.60  & 2.11  & 2.20& 0.50&5.72  &  1700 & \cellcolor{yellow!60}4.00 \\
    
    Ours (16 Points)  & 2.23  & 1.67  & 1.74 &0.40  &\cellcolor{pink!60} 5.16  & 1.12  & 0.75  & 0.78 &\cellcolor{yellow!60} 0.14&  2.79  &1.64  & 0.95  & 1.05 & 0.17& 4.30  &  1700     &\cellcolor{yellow!60}4.00  \\

    Ours (256 Points, w/o depth)  &2.80  & 1.89  & 2.22 &0.57  & 6.39  & 1.42  & 1.02  & 1.08 & 0.33&  3.11  &2.39  & 1.75  & 1.90 & 0.42& 5.50  &   3120   & 7.03  \\
    
    % Ours (4096 Points, w/o xyz)  &/  &/  &/ &/  & / & 0.84  & 0.57  & 0.64& 0.11&  2.06  &1.54  & 0.63 & 0.82 & 0.10& 4.47  &   3120   & 0.00191  \\
    
    % Ours (256 Points, w/o xyz)  &/  &/  &/ &/  & / & 0.89  & 0.55  & 0.63& 0.11&  2.19  &1.13  & 0.54 & 0.67 & 0.12& 3.13  &   3120   & 0.00191  \\

    Ours (256 Points) & \cellcolor{yellow!60}2.20  & \cellcolor{yellow!60}1.42  & \cellcolor{yellow!60}1.58  &\cellcolor{yellow!60}0.34& \cellcolor{yellow!60}5.30  & \cellcolor{yellow!60}0.88  & \cellcolor{yellow!60}0.53  & \cellcolor{yellow!60}0.62  & 0.17&\cellcolor{yellow!60}2.15  & \cellcolor{yellow!60}1.05  & \cellcolor{yellow!60}0.37  &\cellcolor{yellow!60} 0.55 & \cellcolor{yellow!60}0.09& \cellcolor{yellow!60}3.02  &   3120    & 7.03  \\
    
    Ours (4096 Points) & \cellcolor{pink!60}2.18  & \cellcolor{pink!60}1.30  & \cellcolor{pink!60}1.54  & \cellcolor{pink!60}0.33&5.39  &\cellcolor{pink!60}0.78  &\cellcolor{pink!60}0.42  &\cellcolor{pink!60} 0.51  &\cellcolor{pink!60}0.11 &\cellcolor{pink!60}1.97  &\cellcolor{pink!60}0.99  &\cellcolor{pink!60}0.28  &\cellcolor{pink!60}0.50 &\cellcolor{pink!60}0.08 &\cellcolor{pink!60} 2.94  &   5700    & 47.12 \\

    \bottomrule
        % \multicolumn{21}{l}{We report the test time of FFCC from its paper, the training time is with "/" since we have different size of dataset.}
    \end{tabular}
    \begin{tablenotes}    %add table note
        \footnotesize               
        \item[1] Since we have multi-scale datasets, the train time is for the DepthAWB dataset.
        \item[2] The profiled time for PCCC includes the time slot for aligning RGB image and depth image, building point cloud and also network inferring.  
      \end{tablenotes}          
    \end{threeparttable}  
     }%
    \caption{Quantivate comparison of our reannotated NYU-v2~\cite{silberman2012indoor}, Diode~\cite{vasiljevic2019DIODE}, ETH3D~\cite{2017ETH3d} datasets and our collected dataset, respectively. We highlight top two performers in each metric with pink and yellow background, respectively. All test time are CPU time.}
  \label{tab:NYU}%
\end{table*}%

\begin{table*}[h]
  \centering
  \resizebox{\textwidth}{!}{
    \begin{tabular}{lcrr||cccc||cccc||cccc}
    \toprule
    {BackBone} & Image &Train &FPS  & \multicolumn{4}{c||}{NYUv2\&Diode} & \multicolumn{4}{c||}{ETH3d RAW} & \multicolumn{4}{c}{DepthAWB} \\
          & Scale &time &  & Mean  & Med.  & Tri.  & W25\%   & Mean  & Med.  & Tri.  & \multicolumn{1}{c||}{W25\%} & \multicolumn{1}{c}{Mean} & \multicolumn{1}{c}{Med.} & \multicolumn{1}{c}{Tri.} & \multicolumn{1}{c}{W25\%} \\
    \midrule
    PointCNN~\cite{li2018pointcnn} & 16*16 & $\sim 2$ &  $\sim 20$   & 2.41  & 1.68  & 1.83  & 5.58  & 1.55  & 1.33  & 1.35  & 3.09  &  2.59     &1.78       &1.95       &5.92  \\
    PointCNN~\cite{li2018pointcnn} & 64*64 &  $>24$ & $\sim 5$  & \cellcolor{pink!60}2.19  & \cellcolor{yellow!60}1.55  & \cellcolor{yellow!60}1.69  & 5.02  & 1.30      & 1.02      &1.11       & 2.72      &  1.31     & 0.91      &0.97       &3.00  \\
    PointNet++~\cite{qi2017pointnet++} & 16*16 &  $\sim 3$  & $\sim 100$  & 2.69  & 2.87  & 3.76  & \cellcolor{pink!60}{3.59}  & 1.44  & 1.50  & 1.52  & \cellcolor{pink!60}{1.93} & 1.61      &  1.56     & 1.57      &\cellcolor{pink!60}{2.39}  \\
    PointNet++~\cite{qi2017pointnet++} & 64*64 &    $>24$ & $\sim 3$  &2.98       &1.98       & 2.20      & 7.04      &  1.66     & 1.25      & 1.34      & 3.39      & 1.91      & 2.02      &  1.99     &  2.57\\
    Ours  & 16*16 &\cellcolor{pink!60}  $\sim 0.5$   &\cellcolor{pink!60}$\sim 530$  & \cellcolor{yellow!60}2.35  & 1.58  & 1.80  & 5.37  & \cellcolor{yellow!60}1.08  & \cellcolor{yellow!60}0.78  & \cellcolor{yellow!60}0.87  & 2.43  &\cellcolor{yellow!60}1.05      & \cellcolor{yellow!60}0.37      &\cellcolor{yellow!60}0.55      &3.02  \\
    Ours  & 64*64 & \cellcolor{yellow!60} $\sim 1.5$     &\cellcolor{yellow!60}$\sim 495$ & \cellcolor{yellow!60}2.20  & \cellcolor{pink!60}{1.30}  & \cellcolor{pink!60}{1.57}  & \cellcolor{yellow!60}5.53  & \cellcolor{pink!60}{0.87}  & \cellcolor{pink!60}{0.52}  & \cellcolor{pink!60}{0.61}  &\cellcolor{yellow!60}2.13  & \cellcolor{pink!60}{0.99}      & \cellcolor{pink!60}{0.28}      & \cellcolor{pink!60}{0.50}      &\cellcolor{yellow!60}2.94  \\
    \bottomrule
    \end{tabular}}%%
  \caption{Quantitative results of point cloud based backbones. We test three different backbones on three datasets, with two different input scales respectively. Training time in hours. FPS is from GPU. Color setting as in Table \ref{tab:NYU}.}
  \label{tab:backbone}%
\end{table*}%
\section{Experiments}
\label{Sec:Experiments}

We evaluate PCCC and a large list of prior methods on three public RGB-D datasets (relabeled by us, Section \ref{Sec:NYU} and \ref{Sec:ETH}) and a newly-collected RGB-D dataset (Section \ref{Sec:Depth-awb}).  
The dataset overview is as follows: 
\begin{compactitem}
    \item \textbf{NYU-v2 \& Diode Dataset}: 324 linearized RGB images and their corresponding depth map from NYU-v2 \cite{silberman2012indoor} and Diode datasets~\cite{vasiljevic2019DIODE}.
    \item \textbf{ETH3D Dataset}: 195 reprocessed Raw RGB images and their corresponding depth map from ETH-3D datasets~\cite{2017ETH3d}.
    \item \textbf{DepthAWB Dataset}: 185 DSLR RGB images captured by LUMIX GH5S and depth maps by Intel Realsense L515.
\end{compactitem}

\textbf{Training:} We use Pytorch to build and train our method on NVIDIA V100 (Tesla) GPU, Adam \cite{kingma2014adam} is employed as the optimizer. The learning rate is 0.0003. The epoch is set to 10k. 

\textbf{Metrics:} Similar as prior color constancy tasks \cite{xiao2020multi,xu2020end,yu2020aaai}, we report the mean, median (Med.), trimean (Tri.), best (B) and worst (W) 25\% of angular error (Eqution (\ref{Equ:ang})). To validate our method's time efficiency, we also present the time cost during training (for the learning-based methods) and testing (for all).

\renewcommand{\thefootnote}{2}
\subsection{Quantitative Results}
\label{Sec:QR}

% To evaluate our method, we carefully compare it to a dozen of prior illuminant estimation methods, including learning-based and learning-free methods.

Table~\ref{tab:NYU} presents the quantitative results of PCCC and a large selection of prior arts on three RGB-D color constancy datasets. In overall, the proposed PCCC (with 4096 points) leads three leaderboards, obtaining the lowest error in almost all statistics. 
Shrinking the input size to $256$ points\footnote{This equals to a $16\times 16$ RGB image plus a same-size depth map.},  PCCC also performs steadily better than the state-of-the-art learning based methods \cite{barron2017fourier,hu2017cvpr} by a large margin, again showing the advantage of PCCC. 
To our a bit surprise, the PCCC model with  $16$-points input still delivers meaningful results, demonstrating the net learns the illumination from the 96 float values w.r.t. image.

\begin{table}[tbh]
  \centering
  \resizebox{\linewidth}{!}{
    
    \begin{tabular}{ccc||cccc|cccc}
    \toprule
    \multicolumn{3}{c||}{Module}  & \multicolumn{4}{c|}{Angular Error (16x16)}  & \multicolumn{4}{c}{Angular Error (64x64)}\\
    CPA   & LIA   & SW     & Mean  & Med   & Tri.  & W25\%  & Mean  & Med   & Tri.  & W25\% \\
    \midrule
    - & - & - &        1.32           &     0.81  & 0.90      &    3.35   & 1.33      &     0.75  & 0.90      & 3.61       \\
    - & - & \checkmark            &   1.23    &   0.79    &    0.88   &   3.14    &             1.18      &  0.53     &    0.70   &  3.19     \\
    - & \checkmark      & \checkmark            &  \cellcolor{yellow!60}1.14  &    0.57   &  \cellcolor{yellow!60}0.72   &   \cellcolor{yellow!60}3.13   & \cellcolor{yellow!60}1.04    &0.43      & \cellcolor{yellow!60}0.60     &\cellcolor{yellow!60}2.95             \\
    \checkmark &  -     & \checkmark          &1.19       &\cellcolor{yellow!60}0.50       &\cellcolor{yellow!60}0.72       &3.32            &1.10     &\cellcolor{yellow!60}0.42      &0.63       &3.16      \\
    \checkmark      &  \checkmark     &    \checkmark             &\cellcolor{pink!60}1.05  & \cellcolor{pink!60}0.37     &\cellcolor{pink!60}0.55      & \cellcolor{pink!60}3.02  & \cellcolor{pink!60}0.99     &\cellcolor{pink!60}0.28     & \cellcolor{pink!60}0.50     &\cellcolor{pink!60}2.94 \\
    \bottomrule
    \end{tabular}}%
  \caption{PCCC ablation study. On the DepthAWB dataset, for we two resolutions of the point cloud -- 16x16 (left) and 64x64 (right). CPA -  camera pose augmentation, LIA - light intensity augmentation; both described in Section \ref{Sec:Aug}. SW - refers to the spatial weighting, see Section \ref{Sec:Regressor}. Color setting as in Table\ref{tab:NYU}.}
  \label{tab:abblation}%
\end{table}%

%To Yanlin: explain depth importance here 
From a multiview point of view, 3D point cloud can be rendered to 2D images, as done in \cite{qi2016volumetric}, which forms a projection-based augmentation boosting the task performance. 
%We observe that  the accuracy degenerates averagely by $0.30\sim 0.65$ degrees, but is still performs over the majority of learning-based methods on ETH3d and DepthAWB dataset. 
%but weaker than the top-performing  \cite{barron2017fourier,hu2017cvpr} on ETH3d and DepthAWB dataset. 

%\textcolor{blue}{shell we mention the contribution of depth?}

It worth mentioning that all PCCC variants provide an over-realtime inference time, nearly $7$ mini-seconds (equals to $140$ frames per second) on CPU.  Considering the tiny input size,  we believe it is a viable solution for smartphone photography. 

% Table \ref{tab:NYU} shows the quantitative results on the two annotating datasets and our collected dataset, our method outperforms other methods with a clear margin, while using 4 to 16 times smaller input size than state-of-the-art learning based methods \cite{barron2017fourier,hu2017cvpr} and achieving 700fps in test set. 

Figure~\ref{fig:mainshow} shows a qualitative comparison of PCCC and our selected well-performing methods. Among all choices, the proposed method obtains the lowest error and delivers the most neural appearance, for both outdoor and indoor scenarios. Please see supplement for more visual results. 

% Figure \ref{fig:mainshow} illustrates our visual result comparing to recent innovated computational color constancy methods, our method performs best at all separate images, we maintain a stable performance on both indoor and outdoor scenario, our method also works well on images with large pure color. 

\textbf{Notice} All learning-based methods and their hyper-parameters are from their official implementations or carefully tuned. All test time are the CPU time. 

\subsection{What Depth Brings}
In Table~\ref{tab:NYU}, we also report the results when PCCC is not given depth (in practice, it is done by initializing the depth as $1$ uniformly).
We observe that, for 16-points and 256-points models, the accuracy degenerates by a noticeable gap, proving the contribution of the depth. On ETH3d and DepthAWB datasets, the PCCC variant without depth still performs over the majority of learning-based methods, showing that it learns to  infer illumination chromaticity well from a ``flatten'' point cloud. Empirically,  we deduce that depth itself contains color information but provides geometric clue for PCCC to tell how the illumination varies from one location to another in a 3D space. 

\begin{figure}[ht]
    \centering
    \includegraphics[width=0.9\linewidth]{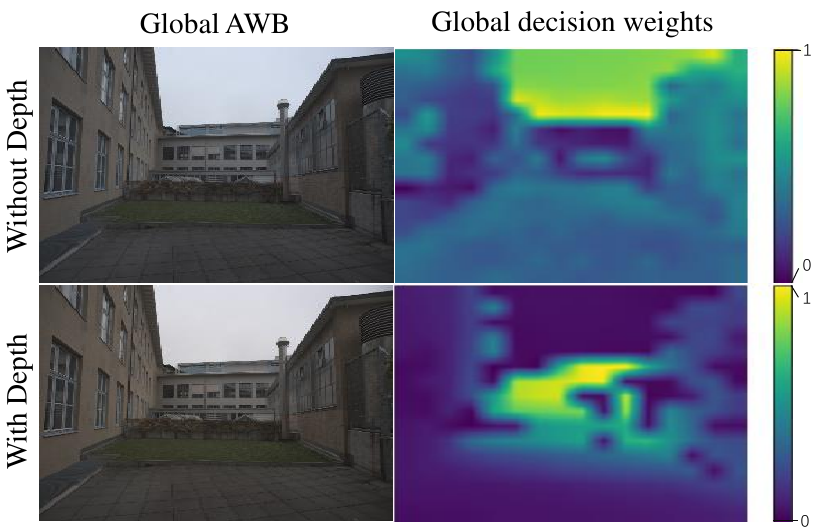}
    \caption{ Illustration of  how the depth affects the final decision. Depth information helps PCCC to estimate illumination from more meaningful information (building wall) instead of noisy information (bright sky with disturbing color).}
    \label{fig:depthuse}
\end{figure}

\begin{figure*}[!h]
    \centering
    \resizebox{\textwidth}{8.5cm}{
    \includegraphics[width=\linewidth]{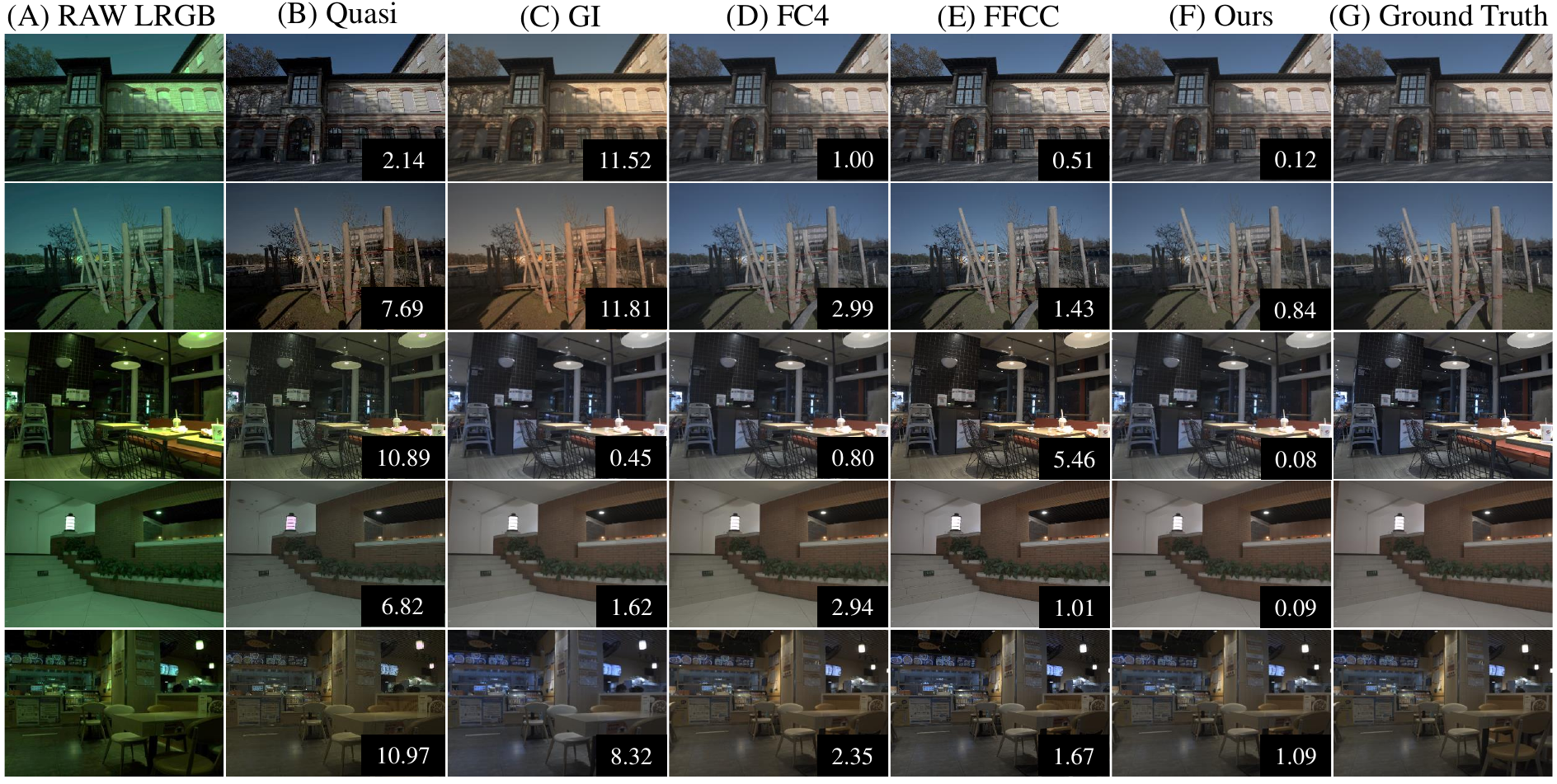}}
    \caption{Quantitative global illuminant estimation results on the relabeled ETH3d dataset~\cite{2017ETH3d} and the new DepthAWB dataset. 
    (A) and (G) are Raw linear RGB images and images corrected by ground truth illuminants, respectively. From (B) to (F): images with estimated illuminants from Quasi \cite{bianco2019quasi}, Gray Index \cite{Qian_2019_CVPR}, FC4 \cite{hu2017cvpr}, FFCC \cite{barron2017fourier}, and PCCC, respectively. }
    \label{fig:mainshow}
\end{figure*}

 Figure \ref{fig:depthuse} visualizes the illumination certainty map to evaluate the importance of depth information. With depth, PCCC can aggregate  feature more from the area which is of more “achromatic” surface and meaningful depth.
% \textcolor{blue}{add visualization of illuminant certainty}

\subsection{Ablation Study}
\textbf{Point Cloud Processing Backbone}
\label{Sec:Why}
As discussed in Section \ref{Sec:Regressor}, there are obviously many selections  for point cloud based net that can be employed as the backbone of the proposed method.  We replace the PointNet backbone by different point cloud nets (e.g., PonitCNN~\cite{li2018pointcnn}, PointNet++~\cite{qi2017pointnet++}) and test their performance in our task.

Table \ref{tab:backbone} proves that given the input of same size, PointNet backbone gives better mean angular error compared to  PointCNN and PointNet++ backbones, while costs much less time ($\sim$ 1/80 times).  As PointNet backbone achieves a good trade-off between accuracy and efficiency, we adopt PointNet as our default backbone.  For fair comparison, only the backbone is switched while other modules are the same.
% we only switch the backbone while keeping other modules the same. 

% illustrates the quantitative comparison with those operators, our method obtains better angular error, while using less time. PointCNN and PointNet++ perform fair results but cost 80 times more time than our method. To be a fair comparison, we only replace the backbone to different operators, while keeping other operations the same. 

% Table \ref{tab:backbone} illustrates the quantitative comparison with those operators, our method obtains better angular error, while using less time. PointCNN and PointNet++ perform fair results but cost 80 times more time than our method. To be a fair comparison, we only replace the backbone to different operators, while keeping other operations the same. 

% Table generated by Excel2LaTeX from sheet 'Sheet3'

\begin{figure}[ht]
    \centering
    \includegraphics[width=\linewidth]{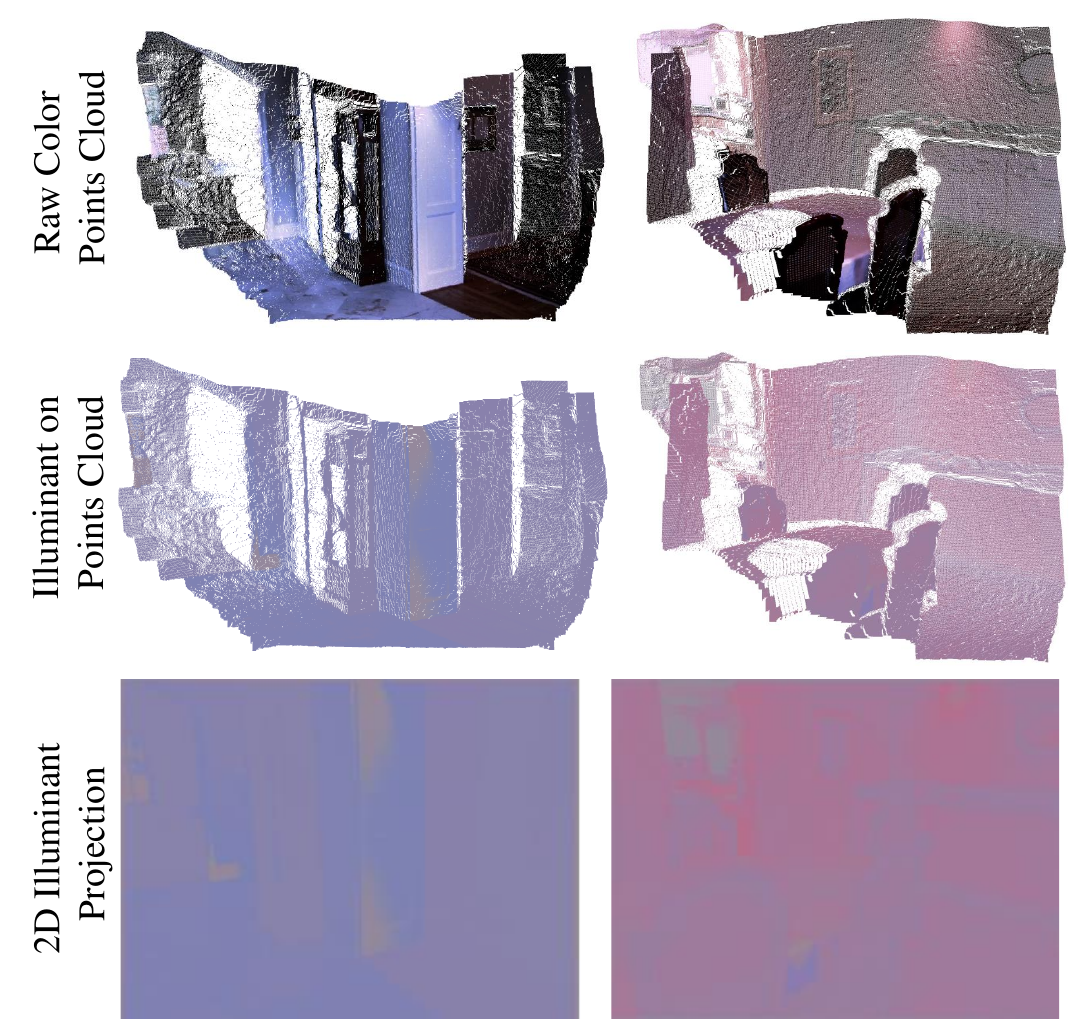}
    \caption{ Illustration of illumination distribution over 3D point clouds and the corresponding 2D illumination maps. }
    \label{fig:localawb}
    \vspace{-3ex}
\end{figure}

\textbf{Point Set Size and Augmentation} 
\label{Sec:Ablation}
We investigate the influence of three proposed modules: camera pose augmentation (CPA), light intensity augmentation (LIA), and spatial weight distribution (SW), using two different size point clouds $16\times16$ and $64\times64$ respectively.

Results are shown in Table \ref{tab:abblation}. Compared to using a very sparsely-sampled point cloud (256 points), the 4096-point point cloud obtains better results on each metric. 
By deactivating each devised module (CPA, LIA, SW) one by one, we notice a gradual performance drop, which validates the benefit of each module. 
%Meanwhile, the two augmentation methods and the spatial weight distribution make contributions to better performances. 

% Delete the dual part or ? and add more experimental details for depth?
% \subsection{Depth Contribution}
% We visualize the illuminant certainty map to evaluate how the depth information help the illumination estimation. 

\subsection{Beyond Global Illumination Estimation}
\label{Sec:Local}

Considering that in PCCC illumination-related feature is deduced point-wisely, it is straight-forward to explore its potential application in point-wise AWB (or referred as local AWB).
% Since our PCCC calculates illumination point-wisely, we can further implement it to point-based operations. We discuss extension abilities of PCCC on Local color constancy and relighting as following. 

% Benefited by point clouds' ... and ... , we can extend our method on spatial illuminant distribution, and relighting tasks.  

%\subsection{Illumination Distribution Over Point Cloud and Local AWB}
% \subsection{Point-wise Illumination}

% \label{Sec:Local}

As introduced in Section \ref{Sec:Regressor}, we can achieve point-wise illumination distribution (later weighted by a weight matrix and fused into the global illuminant). If we take this intermediate feature as our final output, we obtain a $N\times3$ illumination tensor, where $N$ is the point number. 
The illumination tensor can be visulized as a colored point cloud (Figure~\ref{fig:localawb}, 2$^{nd}$ row), and, if projected to pixel coordinate system, as a 2D illumination map (Figure~\ref{fig:localawb}, 3$^{rd}$ row). 

\begin{figure}[!ht]
    \centering
    \includegraphics[width=\linewidth]{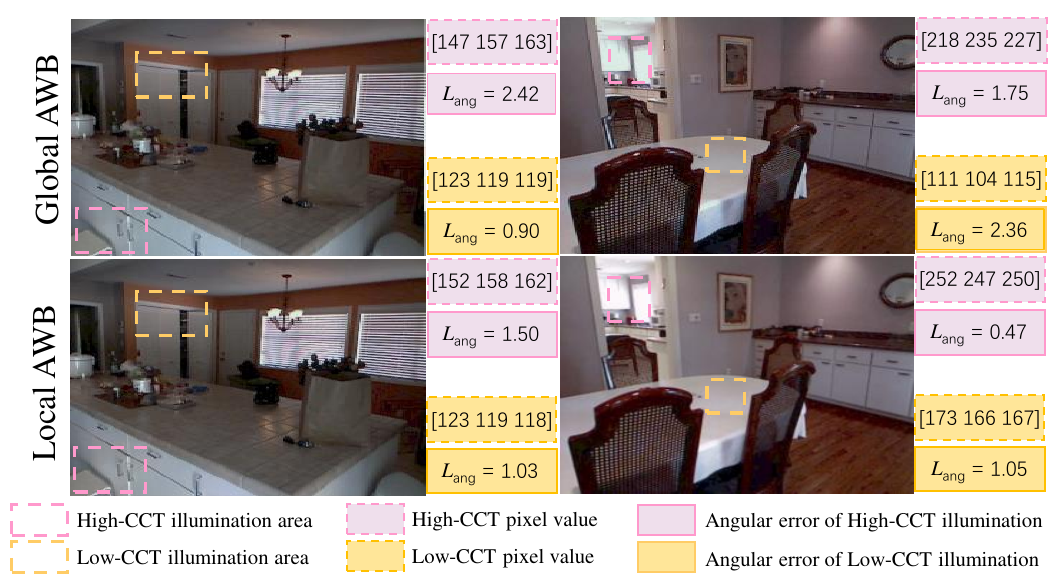}
    \caption{Visualization of Global AWB result and Local AWB result. 
    % To help readers better evaluate the performance of Local AWB, 
    We use yellow and cyan bonding box to indicate the lower and higher correlated color temperature (CCT), respectively. }
    \vspace{-3ex}
    \label{fig:localawbeffect}
\end{figure}

With 2D illumination map, getting the reciprocals for each point's vector yields its AWB gain.  We apply the AWB gain pixelwisely and visualize in Figure~\ref{fig:localawbeffect}. Comparing the image intensities recovered by both solutions,  
we can find the local one more accurately estimate the dual illumination.
% Comparing with Gray Index\cite{Qian_2019_CVPR}, KNN WB \cite{afifi2019color} which also provide local AWB solution, our method better recovers the neutral color in both near-window area and dim indoor room space.  

% \textbf{Limitation} Our local AWB results based on the experimental results in NYU-v2 dataset, which the dataset
% As introduced in Section \ref{Sec:Regressor}, we can achieve point-wise illumination distribution by presenting an implicit weight structure. If we directly use the output from the sptial illuminant $\textbf{P}_{illum}$, we can obtain a $N\times3$ scale tensor, where $N$ equals the input scale of point clouds, and the illuminant tristimulus values are in the three channels respectively. We then concatenate the illuminant tristimulus values to the corresponding position of the original point cloud, and we can obtain the illuminant distribution over the point cloud.  

% Based on it, local AWB can be achieved by applying the point cloud's illuminant distribution.  We present the points cloud wise illumination and its corresbonding 2D projection in Figure \ref{fig:localawb}. The local color constancy comparison can be found in Figure \ref{fig:localawbeffect}, our PCCC recovers best local white balance where both indoor and outdoor illluminnat are estimated and corrected.  

\section{Conclusion}

We develop a point set-based regression net, PCCC, for the color constancy task. By leveraging the geometry information and  extensive point-wise augmentation, it deduces the global illumination chromaticity accurately, even in challenging mixed-lighting environments. On two popular RGB-D datasets which we added illumination information to, and on the novel DepthAWB benchmark, PCCC obtains lower error than over the state-of-the-art,
reaching the   mean angular error equals of $0.99$ degree on DepthAWB.
Its side applications like local AWB is also discussed.  The PCCC method is efficient,
using just $16 \times 16$-size input, and fast, reaching $\sim 140$ fps on CPU. In future work, we would like to study more specific point cloud-based operator for color processing task and its interpretability.  

{\textbf{Limitations:} The quality and quantity of datasets can be further improved, including add Local AWB labels. }

\appendix
\section{Dataset Preprocessing}
Introduced in paper Section 4, we use three public RGBD datasets and collect DepthAWB dataset for point cloud color constancy. We will detail the procedure of labeling the datasets and collecting our dataset.
\subsection{Incorrect White Balance}
\begin{figure}[t]
    \centering
    \includegraphics[width=\linewidth]{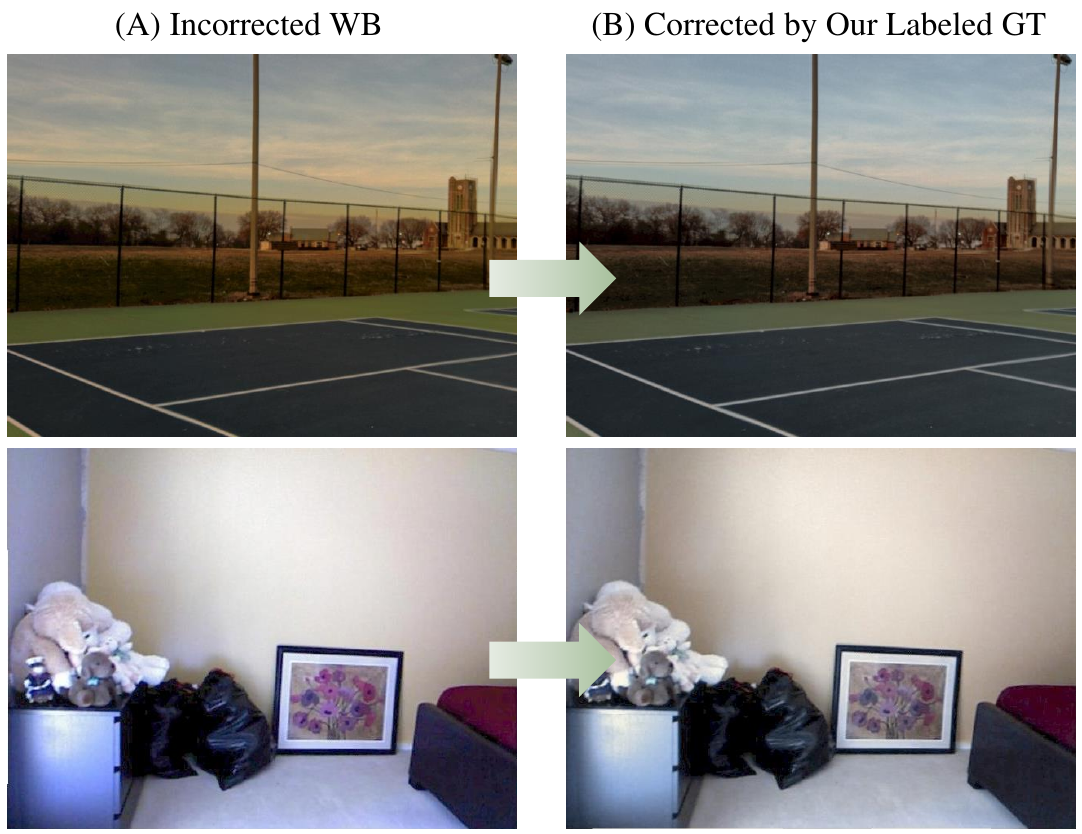}
    \caption{AWB-biased images and their corrected version with our labeled ground-truth illuminations from DIODE datasets\cite{vasiljevic2019DIODE} and NYU-v2 \cite{silberman2012indoor}, respectively.}
    \label{fig:sp_wrongwb}
\end{figure}

In camera ISP auto white balance (AWB) is a fundamental processing step, where the incorrect white balance images may exist due to fragile AWB algorithm or improper manual setting. Figure \ref{fig:sp_wrongwb} shows the AWB-biased samples we found in the NYU-v2 dataset \cite{silberman2012indoor} and DIODE dataset \cite{vasiljevic2019DIODE}, these images are linearized sRGB image.
\subsection{Public Datasets Labeling}

% Label the illumination by finding neutral objects in the scene is a common operation when there is no color checker in datasets. 
 Cheng \etal~\cite{cheng2016cvpr} identified 66 images with two illuminations from Gehler-Shi dataset~\cite{gehler2008bayesian,shi2010re} , and relabeled two illumination vectors by finding neutral objects in the scene. We use the same idea to label the dominant illumination for NYU-v2~\cite{silberman2012indoor}, DIODE~\cite{vasiljevic2019DIODE} and ETH3d~\cite{2017ETH3d} datasets.
Given a raw image $\textbf{I}$, we first locate several areas under the cast of the dominant light, e.g. a piece of printed paper under the sunlight outside the window, and a gray wall illuminated by the ceiling light in the room. Then we visually check whether the area is saturated or underexposed, and select the well-exposed part.
%if the saturation of this area is overexposed, and select the correctly exposed part.
% Finally we calculate the mathematical average of each color based on each pixel in the select area $\textbf{I}_{sel}$ and obtain the global illumination $\textbf{E}$.
As the final step we average the rgb instensities of the selected area  $\textbf{I}_{sel}$, yielding the global illumination $\textbf{E}$: 
\begin{equation}
    \textbf{E} = \frac{1}{N}\sum_{R,G,B}{\textbf{I}_{sel}}.
\end{equation}

\begin{figure}[t]
    \centering
    \includegraphics[width=\linewidth]{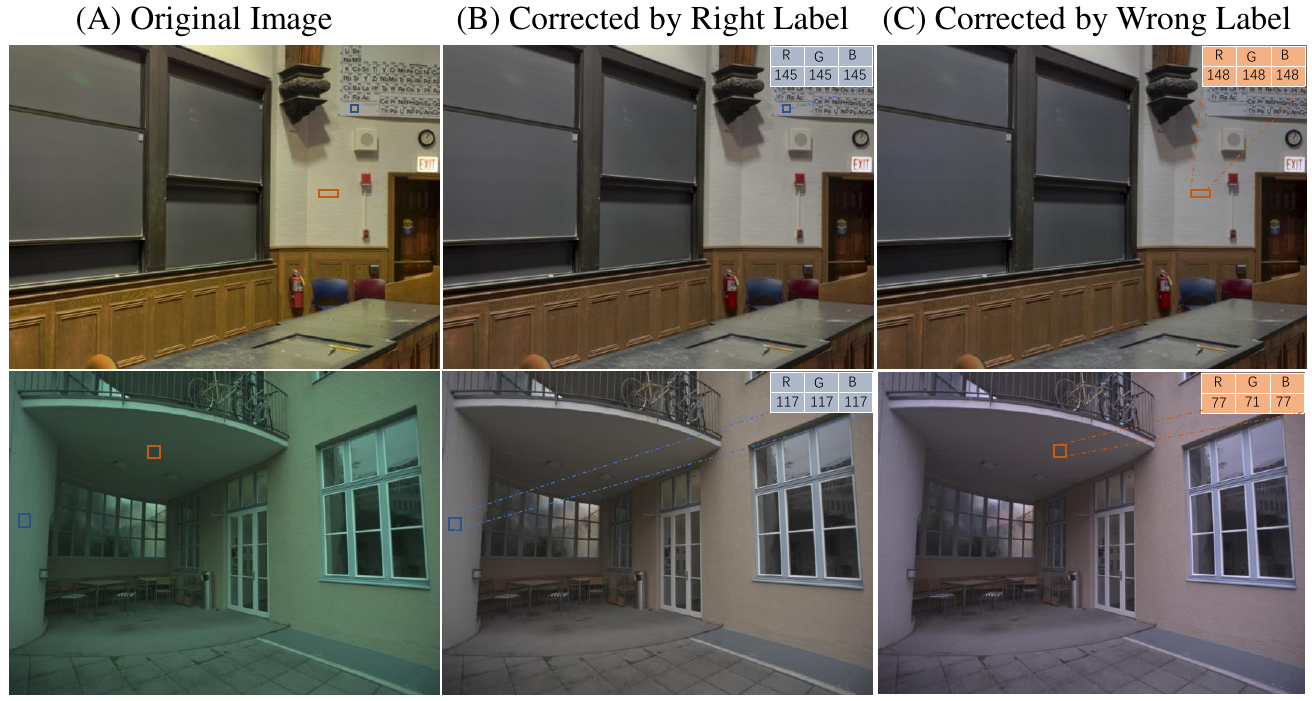}
    \caption{Illustration of the procedure of labeling the illumination by finding canonical neutral surface. We mark the blue and red boxes in column (A) indicating the right and wrong area of ground truth illumination calculation, respectively. Column (B) presents images corrected with right label. Column (C) shows images with wrong label. $\gamma=2.2$ is applied for visualization.}
    \label{fig:sp_label}
\end{figure}

Figure \ref{fig:sp_label} shows the resulting corrected images if we label correctly or with some mistake. 

\subsection{Dataset Collection}
% As introducing in paper Section \ref{Sec:Depth-awb}, we use GH5S as RGB camera and Intel Real-sense L515 LiDAR camera as our depth map collector. 
Our DepthAWB dataset contains of three types data, 1) RGB images by GH5s camera, 2) depth maps by Intel Real-sense L515 LiDAR camera, and 3) illumination ground-truth by labeling from Xrite Color Checker. 

Images from our dataset are mainly collected under indoor scenarios, considering the following reasons: 
1) limited working range of the depth camera, similar as NYU-v2\cite{silberman2012indoor} and SUN RGB-D \cite{song2015sun}, 
and 2) comparing to the outdoor scenarios, the indoor scenes have more artificial and mixed illumination which makes the illumination estimation more challenging.

\begin{figure}[t]
    \centering
    \includegraphics[width=\linewidth]{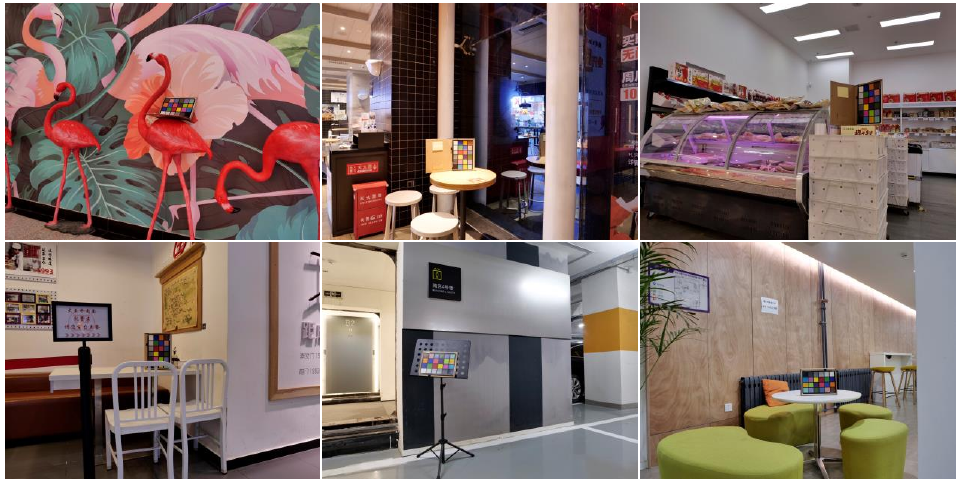}
    \caption{Selected scenes from DepthAWB dataset}
    \label{fig:sp_ourdata}
\end{figure}

% introduce the ideal scenes
We collect our data in several different places to ensure diverse scenes and wide illumination distribution. We are interested in challenging scenes, \eg, shopping mall, large pure color area, and the scene containing illumination-like misleading color. We also pay attention to areas under mixed-illumination but have one dominant illumination. Figure \ref{fig:sp_ourdata} shows several selected scenes from our dataset.

% how we set camera

\begin{figure}[t]
    \centering
    \includegraphics[width=0.63\linewidth]{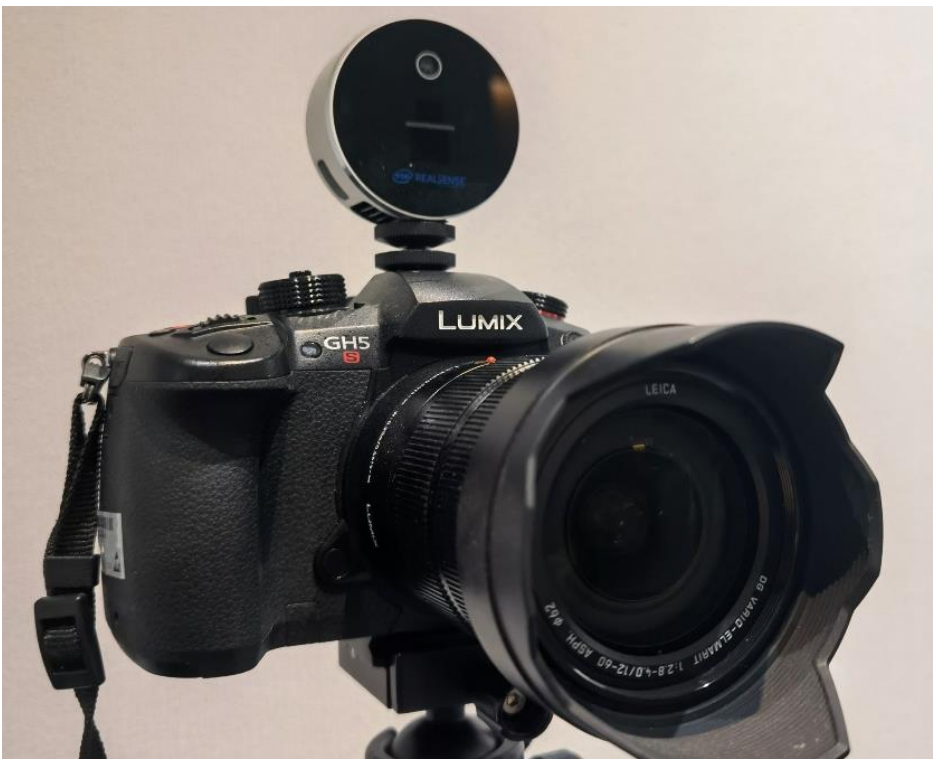}
    \caption{Our data capture setup.}
    \label{fig:setup}
\end{figure}

Our data capture setup is shown in Figure \ref{fig:setup}, the Intel Realsense L515 depth camera is set on the top of Panasonic Lumix GH5s camera.

We fix the camera focal length during shooting, the camera aperture is set as $f/8.0$, and we use aperture priority mode to yield a clear vision and less noise. RGB camera is placed on a tripod and ToF camera keeps the rigid connection with RGB camera.

%introduce how we collect
We create a script to collect RGB images and depth maps from two cameras simultaneously, with a single click on the laptop keyboard. 
%introduce how we put label

A reliable illumination label lays on putting the color checker in the right position. In each scene, we put the color checker under the dominant illumination, and make sure the color checker reflects the dominant illumination into the camera. For the similar scenes under same dominant illumination, we direct use the first labeled illumination. We took two images at one single scene, one image with color checker is for the illumination labeling, and another without color checker is for the training and testing period. 
%introduce the similar illumination

We obtain the ground-truth illumination using the right exposed grayscale color on color checker.

\subsection{Illumination Distribution}
Theoretically, computational color constancy is to learn a mapping pattern from images in wild to illumination in wild. Therefore, a fair and complete illumination distribution is needed. We compare our collection and relabeled datasets to two widely used color constancy datasets on illumination distribution. Figure 4. in main paper shows the illumination distribution of each datasets, similar as the NUS-600D color constancy datasets, the illumination distribution of our datasets are also following the blackbody radiation curve. And Our datasets contain a relatively large range of illumination.
\section{Camera Calibration}
We present the camera intrics we used or calibrated in three different datasets.
\subsection{DepthAWB Dataset}
DepthAWB dataset RGB camera (Under (480,640)):
\begin{equation}
\left[\begin{array}{ccc}
     464.0010&0  &319.4235\\
     0&463.1813  &241.4676\\
     0&0  &1
\end{array}\right]
\end{equation}

DepthAWB dataset RGB camera (Under (240,320)):
\begin{equation}
\left[\begin{array}{ccc}
     232.1495&0  &157.8825\\
     0&232.0269  &123.0501\\
     0&0  &1
\end{array}\right]
\end{equation}

DepthAWB dataset Depth camera (Under (480,640)):
\begin{equation}
\left[\begin{array}{ccc}
     470.2773&0  &295.0742\\
     0&470.2187  &242.5917\\
     0&0  &1
\end{array}\right]
\end{equation}

DepthAWB dataset Depth camera (Under (240,320)):
\begin{equation}
\left[\begin{array}{ccc}
     233.5576&0  &148.4298\\
     0&233.6542  &125.9854\\
     0&0  &1
\end{array}\right]
\end{equation}
Rotation:
\begin{equation}
\left[\begin{array}{ccc}
0.9999&0.0096&-0.0103\\
-0.0097&0.9999&	-0.0063\\
0.0102&0.0064&	0.9999
\end{array}\right]
\end{equation}
Shift:
\begin{equation}
\left[\begin{array}{ccc}
0.8710&-105.6946&-90.1409
\end{array}\right]
\end{equation}
\subsection{NYU-v2 Dataset}
NYU dataset RGB camera (Under (480,640)):
\begin{equation}
\left[\begin{array}{ccc}
     525.0&0  &319.5\\
     0&525.0  &329.5\\
     0&0  &1
\end{array}\right]
\end{equation}
\subsection{DIODE Dataset}
DIODE dataset RGB camera (Under (480,640)):
\begin{equation}
\left[\begin{array}{ccc}
     866.81&0  &512.0\\
     0&927.06  &384.0\\
     0&0  &1
\end{array}\right]
\end{equation}
\subsection{ETH3D Dataset}
ETH3D presents different RGB camera intrics on different scene, where we only use a common setting in our point cloud building, since we only need the relative distance between pixels.

ETH3D dataset RGB camera (Under (4032,6048)):
\begin{equation}
\left[\begin{array}{ccc}
     3406.79& 0  &3040.861\\
     0& 3404.57  &2014.4\\
     0&0  &1
\end{array}\right]
\end{equation}
\section{More Visual Results}
We present the visual results of global and local illumination estimation.
\subsection{Global Illumination Estimation}
\begin{figure*}
    \centering
    \resizebox{\linewidth}{!}{
    \includegraphics{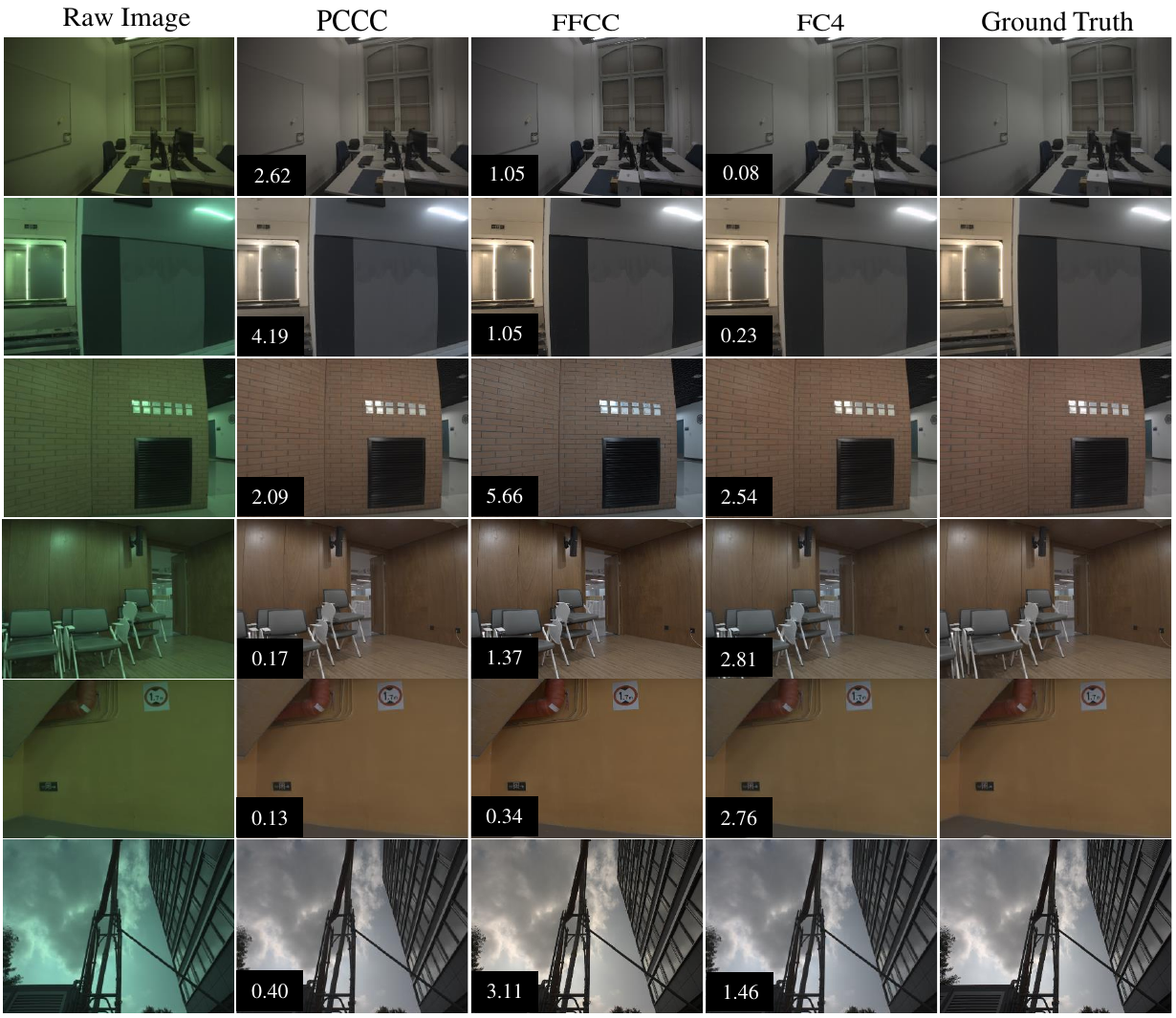}}
    \caption{More visual results of our methods comparing with FC4\cite{hu2017cvpr} and FFCC\cite{barron2017fourier}. The angular error of each image is listed in the black box. We present our failure (angular error $>$ 2) and well (angular error $<$ 0.5) estimation results.}
    \label{fig:spshow}
\end{figure*}

We show more visual comparison results with two state-of-the-art color constancy methods in Figure \ref{fig:spshow}. As introduced in Section 5 of main paper, our PCCC performs well in majority scenes, yet it also has some limitation in solving huge gray-like area (row 1\&2 of Figure \ref{fig:spshow}). Our method outperforms other methods on other scenes, including, dominating pure-color surface (row 3-5 of Figure \ref{fig:spshow}).   
\subsection{Local Illumination Estimation}
Unlike thumbnail size point clouds we used in global illumination estimation task, we feed full-scale point cloud to the learned PCCC to obtain higher resolution illumination map.

\begin{figure*}
    \centering
    \resizebox{\linewidth}{!}{
    \includegraphics{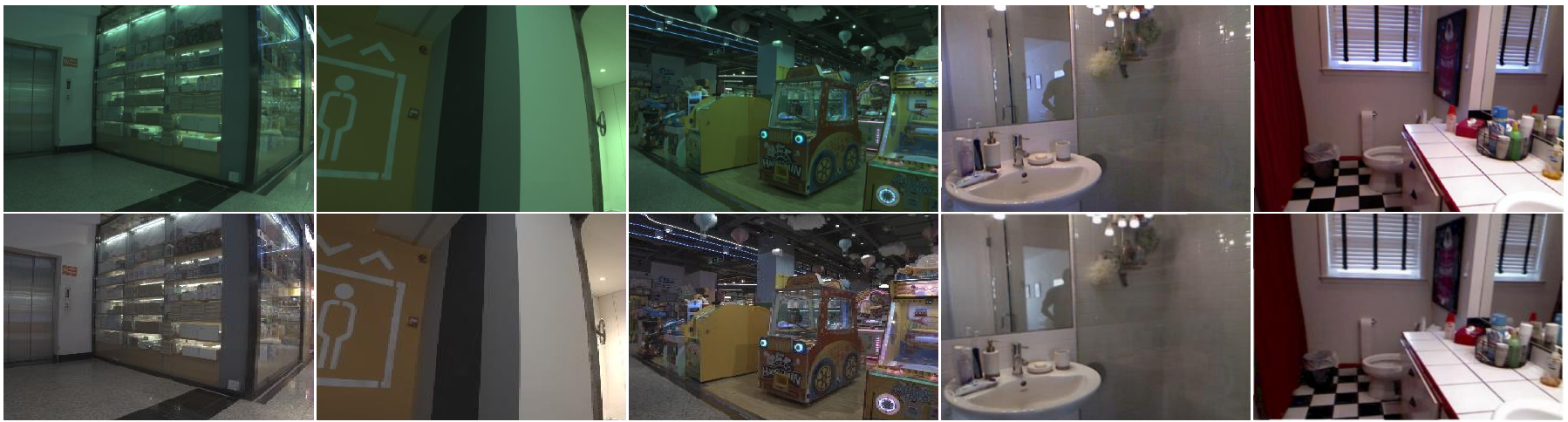}}
    \caption{More visual results of our methods on pixel-wise illumination estimation, we select dual-illuminant scenes from DepthAWB and NYU-v2 datasets, the first row is the original images, and the second row shows images recovered by our pixel-wise illumination map.  }
    \label{fig:sp_dual}
\end{figure*}

Pixel-wise illumination correction results are achieved by the pixel-wise map and presented in Figure \ref{fig:sp_dual}. We select the multi-illuminant scenes from our DepthAWB dataset and NYU-v2 dataset~\cite{silberman2012indoor}. The second row of Figure \ref{fig:sp_dual} shows images recovered by the illumination map. In column 1, our method corrected the illuminations from outside and the windows, separately. In column 4, our method also recovered the outdoor illumination, while keeping the indoor light the same.


{\small
%  \section*{Acknowledgement}
%  This research is supported in part by Shenzhen Stabilization Support Program (No. WDZC20200818121348001).
%  \vspace*{-3ex}
\bibliographystyle{ieee}
\bibliography{egbib}
}

\end{document}